\documentclass[10pt,twocolumn,letterpaper]{article}

\usepackage{cvpr}
\usepackage{times}

\usepackage{amsmath,amssymb,amsthm,amsbsy}
\usepackage{algorithm,algorithmic}
\usepackage{graphicx,subcaption}
\usepackage{booktabs,multirow,diagbox}
\usepackage{enumitem}
\usepackage{cite}
\usepackage{appendix}

\theoremstyle{remark}
\newtheorem{remark}{Remark}

\newcommand{\argmin}{\operatornamewithlimits{argmin}}
\newcommand{\expect}{\mathbb{E}}

\newcommand{\KL}{\mathcal{D}}

\newcommand{\bx}{\mathbf{x}}
\newcommand{\by}{\mathbf{y}}
\newcommand{\bz}{\mathbf{z}}

\newcommand{\bt}{\mathbf{t}}
\newcommand{\bs}{\mathbf{s}}
\newcommand{\bg}{\mathbf{g}}

\newcommand\Mark[1]{\textsuperscript#1}

\usepackage[breaklinks=true,bookmarks=false]{hyperref}
\cvprfinalcopy % *** Uncomment this line for the final submission

\def\cvprPaperID{****} % *** Enter the CVPR Paper ID here

\ifcvprfinal\fi
\begin{document}

%%%%%%%%% TITLE
\title{Data-Free Network Quantization With Adversarial Knowledge Distillation}

\author{Yoojin Choi\Mark{1}, Jihwan Choi\Mark{2}\thanks{Work done when the author was with Samsung as a visiting scholar.} , Mostafa El-Khamy\Mark{1}, Jungwon Lee\Mark{1}\\
\begin{tabular}{cc}
\Mark{1}SoC R\&D, Samsung Semiconductor Inc., San Diego, CA & \Mark{2}DGIST, Korea\\
{\tt\small \{yoojin.c,mostafa.e,jungwon2.lee\}@samsung.com} & {\tt\small jhchoi@dgist.ac.kr}
\end{tabular}
}

\maketitle
\ifcvprfinal\thispagestyle{empty}\fi

%%%%%%%%% ABSTRACT
\begin{abstract}
Network quantization is an essential procedure in deep learning for development of efficient fixed-point inference models on mobile or edge platforms. However, as datasets grow larger and privacy regulations become stricter, data sharing for model compression gets more difficult and restricted. In this paper, we consider data-free network quantization with synthetic data. The synthetic data are generated from a generator, while no data are used in training the generator and in quantization. To this end, we propose data-free adversarial knowledge distillation, which minimizes the maximum distance between the outputs of the teacher and the (quantized) student for any adversarial samples from a generator. To generate adversarial samples similar to the original data, we additionally propose matching statistics from the batch normalization layers for generated data and the original data in the teacher. Furthermore, we show the gain of producing diverse adversarial samples by using multiple generators and multiple students. Our experiments show the state-of-the-art data-free model compression and quantization results for (wide) residual networks and MobileNet on SVHN, CIFAR-10, CIFAR-100, and Tiny-ImageNet datasets. The accuracy losses compared to using the original datasets are shown to be very minimal.
\end{abstract}\vspace{-1em}

\section{Introduction} \label{sec:intro}

\begin{figure}[t]
\centering
\includegraphics[width=.93\columnwidth]{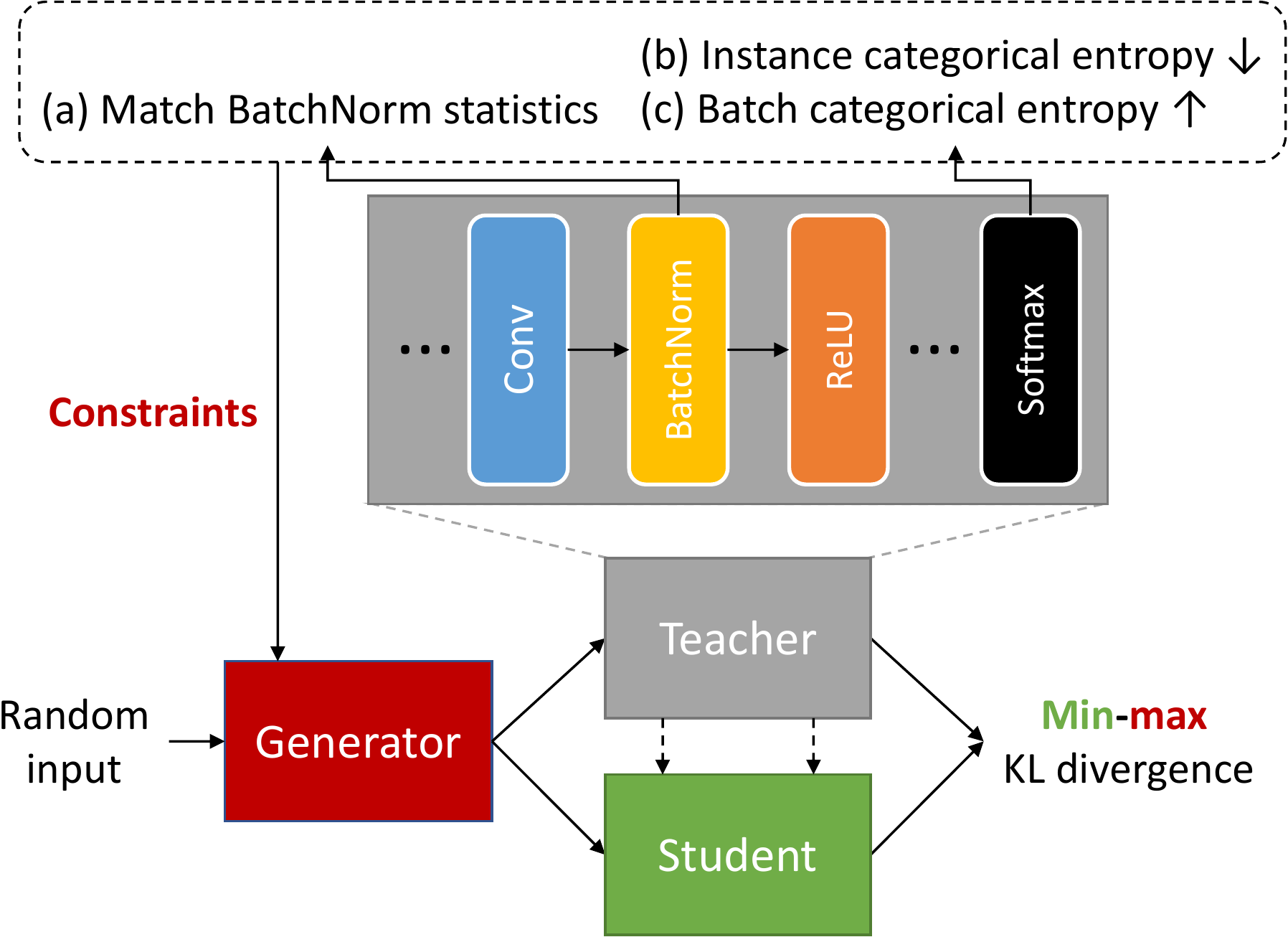}\vspace{-.5em}
\caption{Data-free adversarial knowledge distillation. We minimize the maximum of the Kullback-Leibler (KL) divergence between the teacher and student outputs. In the maximization step for training the generator to produce adversarial images, the generator is constrained to produce synthetic images similar to the original data by matching the statistics from the batch normalization layers of the teacher.
%In the minimization step for KD, intermediate layer outputs can be matched optionally in addition to minimizing the KL divergence between the teacher and student softmax outputs.
\label{sec:intro:fig:01}}\vspace{-1em}
\end{figure}

Deep learning is now leading many performance breakthroughs in various computer vision tasks~\cite{lecun2015deep}. The state-of-the-art performance of deep learning came with over-parameterized deep neural networks, which enable extracting useful representations (features) of the data automatically for a target task, when trained on a very large dataset. The optimization framework of deep neural networks with stochastic gradient descent has become very fast and efficient recently with the backpropagation technique~\cite[Section~6.5]{goodfellow2016deep}, using hardware units specialized for matrix/tensor computations such as graphical processing units (GPUs). The benefit of over-parameterization is empirically shown to be the key factor of the great success of deep learning, but once we find a well-trained high-accuracy model, its deployment on various inference platforms faces different requirements and challenges~\cite{sze2017efficient,cheng2018model}. In particular, to deploy pre-trained models on resource-limited platforms such as mobile or edge devices, computational costs and memory requirements are the critical factors that need to be considered carefully for efficient inference. Hence, model compression, also called network compression, is an important procedure for development of efficient inference models.

Model compression includes various methods such as (1) weight pruning, (2) network quantization, and (3) distillation to a network with a more efficient architecture. Weight pruning and network quantization reduce the computational cost as well as the storage/memory size, without altering the network architecture. Weight pruning compresses a model by removing redundant weights completely from it, i.e., by setting them to be zero, so we can skip computation as well as memorization for the pruned weights~\cite{han2015learning,wen2016learning,guo2016dynamic,molchanov2017variational,louizos2017bayesian,louizos2018learning,frankle2018lottery,dai2018compressing}. Network quantization reduces the memory footprint for weights and activations by quantization and is usually followed by lossless source coding for compression~\cite{han2015deep,choi2017towards,ullrich2017soft,park2017weighted,tung2018deep,choi2020universal}. Moreover, the convolutional and fully-connected layers can be implemented with low-precision fixed-point operations, e.g., 8-bit fixed-point operations, to lower latency and to increase power efficiency~\cite{rastegari2016xnor,zhou2016dorefa,zhu2017trained,cai2017deep,zhang2018lq,jacob2018quantization,wang2019haq}. On the other hand, the network architecture can be modified to be simpler and easier to implement on a target platform. For example, the number of layers and/or the number of channels in each layer can be curtailed. Conventional spatial-domain convolution can be replaced with more efficient depth-wise separable convolution as in MobileNet~\cite{howard2017mobilenets}.
%Inverted residuals and linear bottlenecks in MobileNetV2~\cite{sandler2018mobilenetv2}, group convolution and channel shuffling in ShuffleNet~\cite{zhang2018shufflenet}, and squeeze-and-excitation modules in MobileNetV3~\cite{hu2018squeeze,howard2019searching} are another examples of resource-efficient blocks.

Knowledge distillation (KD) is a well-known knowledge transfer framework to train a small ``student" network under a guidance of a large pre-trained ``teacher'' model. The original idea from Hinton et al. in \cite{hinton2015distilling} utilizes the soft decision output of a well-trained classification model in order to help to train another small-size network. This original idea was further refined and advanced mostly (1) by introducing losses of matching the outputs from intermediate layers of the teacher and student~\cite{romero2015fitnets,zagoruyko2017paying,ahn2019variational}, and (2) by using more sophisticate distance metrics, for example, mutual relations for multiple samples~\cite{chen2018darkrank,park2019relational}.

One issue with existing model compression approaches (including KD) is that they are developed under a strong assumption that the original training data is accessible during the compression procedure. As datasets get larger, the distribution of datasets becomes more expensive and more difficult. Additionally, data privacy and security have emerged as one of primary concerns in deep learning. Consequently, regulations and compliance requirements around security and privacy complicate both data sharing by the original model trainer and data collection by the model compressor, for example, in the case of medical and bio-metric data. Thus, there is a strong need to compress a pre-trained model without access to the original or even alternative datasets.

There have been some attempts to address the problem of data sharing in model compression~\cite{lopes2017data,bhardwaj2019dream,chen2019data,micaelli2019zero}. They aim to perform KD without the original datasets. The early attempts in \cite{lopes2017data,bhardwaj2019dream} circumvent this issue by assuming that some form of compressed and/or partial information on the original training data is provided instead, called meta-data, to protect the privacy and to reduce the size of the data to share. Given a pre-trained model with meta-data, for example, statistics of activation outputs (feature maps) at any intermediate layers, the input is inferred in a backward manner so it matches the statistics in the meta-data. On the other hand, in \cite{chen2019data,micaelli2019zero}, generators are introduced to produce synthetic samples for KD. Chen et al.~\cite{chen2019data} proposed training a generator by using the pre-trained teacher as a fixed discriminator. Micaelli et al.~\cite{micaelli2019zero} used the mismatch between the teacher and the student as an adversarial loss for training a generator to produce adversarial examples for KD. The previous generator-based KD framework in \cite{chen2019data} is rather heuristic, relying on ad-hoc losses. In \cite{micaelli2019zero}, adversarial examples can be any images far different from the original data, which degrade the KD performance.

In this paper, we propose an adversarial knowledge distillation framework, which minimizes the possible loss for a worst case (maximum loss) via adversarial learning, when the loss with the original training data is not accessible. The key difference from \cite{micaelli2019zero} lies in the fact that given any meta-data, we utilize them to constrain a generator in the adversarial learning framework. To avoid additional efforts to craft new meta-data to share, we use the statistics stored in batch normalization layers to constrain a generator to produce synthetic samples that mimic the original training data. Furthermore, we propose producing diverse synthetic samples by using multiple generators. We also empirically show that performing adversarial KD concurrently for multiple students yields better results. The proposed data-free adversarial KD framework is summarized in Figure~\ref{sec:intro:fig:01}.

For model compression, we perform experiments on two scenarios, (1) data-free KD and (2) data-free network quantization. The proposed scheme shows the state-of-the-art data-free KD performance on residual networks~\cite{he2016deep} and wide residual networks~\cite{zagoruyko2016wide} for SVHN~\cite{netzer2011reading}, CIFAR-10, CIFAR-100~\cite{krizhevsky2009learning}, and Tiny-ImageNet\footnote{\url{https://tiny-imagenet.herokuapp.com}}, compared to the previous work~\cite{chen2019data,micaelli2019zero,yin2019dreaming}. Data-free network quantization (data-free quantization-aware training) has not been investigated before to the best of our knowledge. We use TensorFlow's quantization-aware training~\cite{jacob2018quantization,krishnamoorthi2018quantizing} as the baseline scheme, and we evaluate the performance on residual networks, wide residual networks, and MobileNet trained on various datasets, when quantization-aware training is performed with the synthetic data generated from our data-free KD framework. The experimental results show marginal performance loss from the proposed data-free framework, compared to the case of using the original training datasets.

\section{Related work} \label{sec:related}

%\textbf{Knowledge distillation}. Knowledge distillation was introduced by Hinton et al. in \cite{hinton2015distilling} and has been investigated extensively for knowledge transfer and model compression~\cite{romero2015fitnets,zagoruyko2017paying,ahn2019variational,park2019relational}. Transferring knowledge between intermediate layer features was popular to improve the original idea of matching the final softmax output~\cite{romero2015fitnets,zagoruyko2017paying,ahn2019variational}. On the other hand, transferring knowledge of distance-wise and angle-wise relations among multiple data is also proposed in \cite{park2019relational}.

\textbf{Data-free KD and quantization}. Data-free KD attracts the interest with the need to compress pre-trained models for deployment on resource-limited mobile or edge platforms, while sharing original training data is often restricted due to privacy and license issues. 

Some of early attempts to address this issue suggest using meta-data that are the statistics of intermediate features collected from a pre-trained model in \cite{lopes2017data,bhardwaj2019dream}. For example, the mean and variance of activation outputs for selected intermediate layers are proposed to be collected and assumed to be provided, instead of the original dataset. Given any meta-data, they find samples that help to train student networks by directly inferring them in the image domain such that they produce similar statistics as the meta-data when fed to the teacher. Recent approaches, however, aim to solve this problem without meta-data specifically designed for the data-free KD task. In \cite{nayak2019zero}, class similarities are computed from the weights of the last fully-connected layer, and they are used instead of meta-data. Very recently, it is proposed to use the statistics stored in batch normalization layers with no additional costs instead of crafting new meta-data~\cite{yin2019dreaming}.

On the other hand, some of the previous approaches introduce another network, called generator, that yields synthetic samples for training student networks~\cite{chen2019data,micaelli2019zero,yoo2019knowledge}. They basically propose optimizing a generator so that the generator output produces high accuracy when fed to a pre-trained teacher. Adversarial learning was introduced to produce dynamic samples for which the teacher and the student poorly matched in their classification output and to perform KD on those adversarial samples~\cite{micaelli2019zero}.

To our knowledge, there are few works on data-free network quantization. Weight equalization and bias correction are proposed for data-free weight quantization in \cite{nagel2019data}, but data-free activation quantization is not considered. Weight equalization is a procedure to transform a pre-trained model into a quantization-friendly model by re-distributing (equalizing) its weights across layers so they have smaller deviation in each layer and smaller quantization errors. The biases introduced in activations owing to weight quantization are calculated and corrected with no data but based on the statistics stored in batch normalization layers. We note that no synthetic data are produced in \cite{nagel2019data}, and no data-free quantization-aware training is considered in \cite{nagel2019data}. We compare data-free KD and quantization schemes in Table~\ref{sec:related:tbl:01}.

\setlength{\tabcolsep}{0.4em}
\begin{table}[t]
\centering
\caption{Comparison of data-free KD and network quantization schemes based on (1) how they generate synthetic data and (2) whether they rely on meta-data or not.
%Note that \cite{nagel2019data}, \cite{yin2019dreaming}, and ours use batch normalization statistics, which come with pre-trained teacher models at no additional costs.
\label{sec:related:tbl:01}}\vspace{-.5em}
{\small
\begin{tabular}{c|cc}
\toprule
Synthetic data
 & Meta-data
 & Data-free \\
\hline
Not used
 & N/A
 & \cite{nagel2019data}* \\
Inferred in the image domain
 & \cite{lopes2017data}, \cite{bhardwaj2019dream}
 & \cite{nayak2019zero}, \cite{yin2019dreaming}* \\
Generated from generators
 & N/A
 & \cite{chen2019data}, \cite{micaelli2019zero}, Ours* \\
\bottomrule
\multicolumn{3}{r}{* Used the statistics stored in batch normalization layers.}
\end{tabular}\vspace{-1em}
}
\end{table}

\textbf{Robust optimization}. Robust optimization is a sub-field of optimization that addresses data uncertainty in optimization problems (e.g., see \cite{ben2009robust,bertsimas2011theory}). Under this framework, the objective and constraint functions are assumed to belong to certain sets, called ``uncertainty sets.'' The goal is to make a decision that is feasible no matter what the constraints turn out to be, and optimal for the worst-case objective function. With no data provided, we formulate the problem of data-free KD into a robust optimization problem, while the uncertainty sets are decided based on the pre-trained teacher using the statistics at its batch normalization layers.

\textbf{Adversarial attacks}. Generating synthetic data that fool a pre-trained model is closely related to the problem of adversarial attacks (e.g., see~\cite{akhtar2018threat}). Although their purpose is completely different from ours, the way of generating synthetic data (or adversarial samples) follows a similar procedure. In adversarial attacks, there are also two approaches, i.e., (1) generating adversarial images directly in the image domain~\cite{goodfellow2015explaining,carlini2017towards,madry2018towards} and (2) using generators to produce adversarial images~\cite{poursaeed2018generative,wang2019direct,jang2019adversarial}.

\textbf{Deep image prior}. We also note that generator networks consisting of a series of convolutional layers can be used as a good regularizer that we can impose for image generation as prior~\cite{ulyanov2018deep}. Hence, we adopt generators, instead of adding any prior regularization~\cite{mordvintsev2015inceptionism} that is employed in \cite{yin2019dreaming} to obtain synthetic images without generators.

\textbf{Generative adversarial networks (GANs)}. Adversarial learning is also well-known in GANs~\cite{goodfellow2014generative}. GANs are of great interest in deep learning for image synthesis problems. Mode collapse is one of well-known issues in GANs (e.g., see \cite{goodfellow2016nips}). A straightforward but effective way to overcome mode collapse is to introduce multiple generators and/or multiple discriminators~\cite{durugkar2017generative,nguyen2017dual,arora2017generalization,hoang2018mgan}. We also found that using multiple generators and/or multiple students (a student acts as a discriminator in our case) helps to produce diverse samples and avoid over-fitting in our data-free KD framework.

\section{Data-free model compression} \label{sec:dfmc}

\subsection{Knowledge distillation (KD)} \label{sec:dfmc:kd}

Let $\bt_\theta$ be a general non-linear neural network for classification, which is designed to yield a categorical probability distribution~$P_\theta(y|\bx)$ for the label~$y$ of input $\bx$ over the label set $\mathcal{C}$, i.e., $\bt_\theta(\bx)=[P_\theta(y|\bx)]_{y\in\mathcal{C}}$. Let $\by$ be the one-hot encoded ground-truth label~$y$ over the set $\mathcal{C}$ for input $\bx$. The network~$\bt_\theta$ is pre-trained with a labeled dataset, called training dataset, of probability distribution $p(\bx,\by)$, as below:
\[
\theta^*=\argmin_\theta\expect_{p(\bx,\by)}[\KL(\by,\bt_\theta(\bx))],
\]
where $\expect_{p(\bx,\by)}$ is, in practice, an empirical expectation over the training dataset, and $\KL$ stands for Kullback-Leibler (KL) divergence (e.g., see \cite[Section~2.3]{cover2012elements}); note that the minimization of KL divergence is equivalent to the minimization of cross-entropy, given the distribution~$p(\bx,\by)$.

Suppose that we want to train another neural network~$\bs_\phi$, called ``student'', possibly smaller and less complex than the pre-trained network $\bt_{\theta^*}$, called ``teacher.'' The student also produces its estimate of the categorical probability distribution for input~$\bx$ such that $\bs_\phi(\bx)=[Q_\phi(y|\bx)]_{y\in\mathcal{C}}$. Knowledge distillation~\cite{hinton2015distilling} suggests to optimize the student by
\begin{equation} \label{sec:dfmc:kd:01}
\min_\phi
\expect_{p(\bx,\by)}\left[\KL(\by,\bs_\phi(\bx))
+\lambda\KL(\bt_{\theta^*}(\bx),\bs_\phi(\bx))\right],
\end{equation}
where $\lambda\geq0$; note that we omitted the temperature parameter for simplicity, which can be applied before softmax for $\bt_{\theta^*}$ and $\bs_\phi$ in the second KL divergence term of \eqref{sec:dfmc:kd:01}.

\subsection{Data-free adversarial KD} \label{sec:dfmc:minimaxkd}

As shown in \eqref{sec:dfmc:kd:01}, the original KD is developed under the assumption that a training dataset is given for the expectation over $p(\bx,\by)$. However, sharing a large dataset is expensive and sometimes not even possible due to privacy and security concerns. Hence, it is of interest to devise a method of KD in the situation where the training dataset is not accessible, but only a pre-trained teacher is given.

Robust optimization (e.g. see \cite{ben2009robust}) suggests minimizing the possible loss for a worst case scenario (maximum loss) with adversarial learning under data uncertainty, which is similar to the situation we encounter when we are not given a training dataset for optimization. To adopt the robust minimax optimization (also known as adversarial learning) in KD, we first introduce a generator network~$\bg_\psi$, which is used to produce synthetic adversarial data for the input to KD. Then, using the minimax approach, we propose data-free adversarial KD, which is given by
\begin{equation}\label{sec:dfmc:minimaxkd:02}
\min_\phi\max_{\psi}\{\expect_{p(\bz)}[\KL(\bt_{\theta^*}(\bg_\psi(\bz)),\bs_\phi(\bg_\psi(\bz)))]-\alpha L_\psi\},
\end{equation}
for $\alpha\geq0$, where $L_\psi$ is an additional loss that a pre-trained teacher can provide for the generator based on the generator output. We defer our proposed terms in $L_\psi$ to Section~\ref{sec:dfmc:const}.

\begin{remark}
Comparing \eqref{sec:dfmc:minimaxkd:02} to the original KD in \eqref{sec:dfmc:kd:01}, we omit the first KL divergence term related to ground truth labels:
\begin{equation}\label{sec:dfmc:minimaxkd:03}
\min_\phi\expect_{p(\bx)}[\KL(\bt_{\theta^*}(\bx),\bs_\phi(\bx))].
\end{equation}
If we have a generator~$\bg_{\psi^*}$ optimized to mimic the training data exactly such that $p(\bx)=\int p(\bz)\delta(\bx-\bg_{\psi^*}(\bz))d\bz$, then \eqref{sec:dfmc:minimaxkd:03} reduces to
\[
\min_\phi\expect_{p(\bz)}[\KL(\bt_{\theta^*}(\bg_{\psi^*}(\bz)),\bs_\phi(\bg_{\psi^*}(\bz)))].
\]
However, we do not have access to the original training data and cannot find the optimal generator~$\bg_{\psi^*}$. Instead, we minimize the upper bound of $\expect_{p(\bz)}[\KL(\bt_{\theta^*},\bs_\phi)]$ by solving the minimax problem in \eqref{sec:dfmc:minimaxkd:02}, while we give the generator some constraints with the auxiliary loss~$L_\psi$ for the generator to produce similar data as the original training data.
\end{remark}

\subsection{Generator constraints} \label{sec:dfmc:const}

We consider the following three auxiliary loss terms for the generator in the maximization step of \eqref{sec:dfmc:minimaxkd:02} to make the generator produce ``good'' adversarial samples similar to the original data as much as possible based on the teacher.
\begin{enumerate}[noitemsep,topsep=0em]%[noitemsep,topsep=0em,leftmargin=1.2em]
\item[(a)] \textbf{Batch normalization statistics}. Batch normalization layers contain the mean and variance of layer inputs, which we can utilize as a proxy to confirm that the generator output is similar to the original training data. We propose using the KL divergence of two Gaussian distributions to match the mean and variance stored in batch normalization layers (which are obtained from the original data) and the empirical statistics obtained with the generator output.
\item[(b)] \textbf{Instance categorical entropy}. If the teacher is trained well enough for accurate classification, the generator output is of interest only when the categorical distribution output, i.e., softmax output, of the teacher yields small entropy (the probability for one category should be high); the entropy is minimized to zero if one category has probability $1$. That is, we need small entropy for $\bt_{\theta^*}(\bg_\psi(\bz))$ on each sampled~$\bz$.
\item[(c)] \textbf{Batch categorical entropy}. Assuming that each class appears in the dataset with similar probability, the categorical probability distribution averaged for any batch should tend to uniform distribution where the entropy is maximized to $\log_2|\mathcal{C}|$. That is, we need high entropy for $\expect_{p(\bz)}[\bt_{\theta^*}(\bg_\psi(\bz))]$.
\end{enumerate}

Let $\mu(l,c)$ and $\sigma^2(l,c)$ be the mean and the variance stored in batch normalization layer~$l$ for channel~$c$, which is learned from the original training data. Let $\hat{\mu}_\psi(l,c)$ and $\hat{\sigma}_\psi^2(l,c)$ be the corresponding mean and variance computed for the synthetic samples from the generator~$\bg_\psi$. The auxiliary loss~$L_\psi$ for the generator in \eqref{sec:dfmc:minimaxkd:02} is given by
\begin{multline}\label{sec:dfmc:const:01}
L_\psi
=\sum_{l,c}\KL_{\mathcal{N}}((\hat{\mu}_\psi(l,c),\hat{\sigma}_\psi^2(l,c)),(\mu(l,c),\sigma^2(l,c))) \\
+\expect_{p(\bz)}[H(\bt_{\theta^*}(\bg_\psi(\bz)))]-H(\expect_{p(\bz)}[\bt_{\theta^*}(\bg_\psi(\bz))]),
\end{multline}
where $H$ denotes entropy (e.g., see \cite[Section~2.1]{cover2012elements}), and $\KL_{\mathcal{N}}((\hat{\mu},\hat{\sigma}^2),(\mu,\sigma^2))$ is the KL divergence of two Gaussian distributions, which can be represented as
\begin{equation}\label{sec:dfmc:const:02}
\KL_{\mathcal{N}}((\hat{\mu},\hat{\sigma}^2),(\mu,\sigma^2)) \\
=\frac{(\hat{\mu}-\mu)^2+\hat{\sigma}^2}{2\sigma^2}-\log\frac{\hat{\sigma}}{\sigma}-\frac{1}{2}.
\end{equation}
%In \eqref{sec:dfmc:const:01}, to be more general, one can give different weighting for each term, but we simply use equal weighting here.
\begin{remark}
If $\alpha=0$ in \eqref{sec:dfmc:minimaxkd:02}, the proposed scheme reduces to the adversarial belief matching presented in \cite{micaelli2019zero}. Adding the auxiliary loss~$L_\psi$, we constrain the generator so it produces synthetic images that yield similar statistics in the teacher as the original data, which helps the minimax optimization avoid any adversarial samples that are very different from the original data and leads to better distillation performance (basically we reduce the loss due to fitting the model for ``bad'' examples not close to the original dataset). For (b) and (c), we found that similar entropy loss terms are already proposed in \cite{chen2019data}. Batch normalization statistics are used in \cite{nagel2019data,yin2019dreaming}. Yin et al.~\cite{yin2019dreaming} find synthetic samples directly in the image domain with no generators by optimizing an input batch such that it produces similar batch normalization statistics in a pre-trained model. In contrast, we utilize batch normalization statistics to constrain generators. %to produce synthetic images similar to the original training data.
Furthermore, to match the mean and variance, the squared L2 distance is used in \cite{yin2019dreaming}, while we propose using the KL divergence of two Gaussian distributions, which is a distance measure normalized by scale (i.e., standard deviation~$\sigma$ in \eqref{sec:dfmc:const:02}). In \cite{nagel2019data}, batch normalization statistics are used to calculate any quantization biases for correction. No synthetic images are produced in \cite{nagel2019data}.
\end{remark}

\subsection{Multiple generators and multiple students} \label{sec:dfmc:mult}

Using mixture of generators has been proposed to avoid the mode collapse issue and to yield diverse samples that cover the whole support of a target dataset~\cite{hoang2018mgan}. Similarly we propose training multiple generators in our data-free KD framework to increase the diversity of generated samples.
%We also found that using multiple students in KD improve the performance.
Moreover, using multiple discriminators has been also proposed to reduce the mode collapse problem in GANs~\cite{durugkar2017generative}. A similar idea can be adopted in our framework, since we utilize the KL divergence of the student and teacher outputs as the discriminator output. The average KL divergence between the teacher and the students are maximized in minimax optimization. Intuitively, taking average not only reduces the noise in minimax optimization using stochastic gradient descent, but also steers a generator to produce better adversarial samples that are poorly matched to every student in average. The final objective with multiple generators and multiple students is given by
\[
\min_{\phi_i,1\leq i\leq S}\max_{\psi_j,1\leq j\leq G}\sum_{j=1}^G\left(\frac{1}{S}\sum_{i=1}^S\KL_{\phi_i,\psi_j}-\alpha L_{\psi_j}\right),
\]
\[
\KL_{\phi_i,\psi_j}\triangleq\expect_{p(\bz)}[\KL(\bt_{\theta^*}(\bg_{\psi_j}(\bz)),\bs_{\phi_i}(\bg_{\psi_j}(\bz)))],
\]
where $\bs_{\phi_i}$ is the $i$-th student and $\bg_{\psi_j}$ is the $j$-th generator for $1\leq i\leq S$ and $1\leq j\leq G$.

\subsection{Implementation} \label{sec:imple}

We summarize the proposed data-free adversarial KD scheme in Algorithm~\ref{sec:imple:alg:01}. Let $\bz_1^B$ be the random input batch of size $B$ to generators, and let $\KL_{\phi_i,\psi_j}(\bz_1^B)$ and $L_{\psi_j}(\bz_1^B)$ be the losses computed and averaged over batch~$\bz_1^B$. We suggest ``warm-up'' training of generators, optionally, before the main adversarial KD. In the warm-up stage, we train generators only to minimize the auxiliary loss~$L_{\psi}$ so its output matches batch normalization statistics and entropy constraints when fed to the teacher. This pre-training procedure reduces generation of unreliable samples in the early steps of data-free KD. Furthermore, updating students more frequently than generators reduces the chances of falling into any local maximum in the minimax optimization. In the minimization step, one can additionally match intermediate layer outputs as proposed in \cite{romero2015fitnets,zagoruyko2017paying,ahn2019variational}. Finally, data-free network quantization is implemented by letting the student be a quantized version of the teacher (see Section~\ref{sec:exp:mq}).
%In our experiments, we found that it helps to find a better local minimum with faster convergence in some cases.
%Finally, we use cosine annealing~\cite{loshchilov2016sgdr} for the learning rate in both minimization and maximization steps.

\begin{algorithm}[t]
\caption{Data-free adversarial knowledge distillation.}\label{sec:imple:alg:01}
{\small
\begin{algorithmic}
%\STATE Batch size: $B$
%\STATE Warm-up iterations: $N_{\text{warm-up}}$
%\STATE Adversarial KD iterations: $N$
\STATE Generator update interval: $m\geq1$
%\STATE Pre-trained teacher: $\bt_{\theta^*}$
%\STATE Initialize student(s): $\bs_{\phi_i},1\leq i\leq S$
%\STATE Initialize generator(s): $\bg_{\psi_j},1\leq j\leq G$
\STATE Warm-up training for generators (optional)
\FOR{$n:1$ \textbf{to} $N_{\text{warm-up}}$}
\FOR{$j:1$ \textbf{to} $G$}
\STATE $\bz_1^B\leftarrow[\mathcal{N}(0,I)]_1^B$
%\STATE $L\leftarrow\alpha L_{\psi_j}(\bz_1^B)$
%\STATE $\psi_j\leftarrow\psi_j-\eta\nabla_{\psi_j}L$
\STATE $\psi_j\leftarrow\psi_j-\eta\nabla_{\psi_j}L_{\psi_j}(\bz_1^B)$
\ENDFOR
\ENDFOR
\STATE Adversarial knowledge distillation
\FOR{$n:1$ \textbf{to} $N$}
\STATE Maximization
\IF{$n\equiv0$ \text{mod} $m$}
\FOR{$j:1$ \textbf{to} $G$}
\STATE $\bz_1^B\leftarrow[\mathcal{N}(0,I)]_1^B$
\FOR{$i:1$ \textbf{to} $S$}
\STATE $\KL_{\phi_i,\psi_j}(\bz_1^B)\leftarrow\KL(\bt_{\theta^*}(\bg_{\psi_j}(\bz_1^B)),\bs_{\phi_i}(\bg_{\psi_j}(\bz_1^B)))$
\ENDFOR
%\STATE $L\leftarrow\frac{1}{S}\sum_{i=1}^S\KL_{\phi_i,\psi_j}(\bz_1^B)-\alpha L_{\psi_j}(\bz_1^B)$
%\STATE $\psi_j\leftarrow\psi_j+\eta\nabla_{\psi_j}L$
\STATE $\psi_j\leftarrow\psi_j+\eta\nabla_{\psi_j}(\frac{1}{S}\sum_{i=1}^S\KL_{\phi_i,\psi_j}(\bz_1^B)-\alpha L_{\psi_j}(\bz_1^B))$
\ENDFOR
\ENDIF
\STATE Minimization
\STATE $b\leftarrow\lfloor B/G\rfloor$
\FOR{$j:1$ \textbf{to} $G$}
\STATE $\bz_1^b\leftarrow[\mathcal{N}(0,I)]_1^b$
\STATE $\bx_1^{bj}\leftarrow\text{concatenate}(\bx_1^{b(j-1)},\bg_{\psi_j}(\bz_1^b))$
\ENDFOR
\FOR{$i:1$ \textbf{to} $S$}
%\STATE $L\leftarrow\KL(\bt_{\theta^*}(\bx_1^{bG}),\bs_{\phi_i}(\bx_1^{bG}))$
%\STATE $\phi_i\leftarrow\phi_i-\eta\nabla_{\phi_i}L$
\STATE $\phi_i\leftarrow\phi_i-\eta\nabla_{\phi_i}\KL(\bt_{\theta^*}(\bx_1^{bG}),\bs_{\phi_i}(\bx_1^{bG}))$
\ENDFOR
\ENDFOR
\end{algorithmic}
}
\end{algorithm}

\section{Experiments} \label{sec:exp}

We evaluate the proposed data-free adversarial KD algorithm on two model compression tasks: (1) data-free KD to smaller networks and (2) data-free network quantization.
%We implement the proposed framework with Keras\footnote{\url{https://keras.io}} using TensorFlow 1.15.

\textbf{Generator architecture}. Let \texttt{conv3-$k$} denote a convolutional layer with $k$ $3\times3$ filters and stride $1\times1$. Let \texttt{fc-$k$} be a fully-connected layer with $k$ units. Let \texttt{upsampling} be a $2\times2$ nearest-neighbor upsampling layer. The generator input~$\bz$ is of size $512$ and is sampled from the standard normal distribution. Given that the image size of the original data is \texttt{(W,H,3)}, we build a generator as below:

\begin{table}[h]
\centering
{\scriptsize
\vspace{-1em}\begin{tabular}{c}
\toprule
\texttt{fc-8WH}, \texttt{reshape-(W/8,H/8,512)} \\
\texttt{upsampling}, \texttt{conv3-256}, \texttt{batchnorm}, \texttt{ReLU} \\
\texttt{upsampling}, \texttt{conv3-128}, \texttt{batchnorm}, \texttt{ReLU} \\
\texttt{upsampling}, \texttt{conv3-64},  \texttt{batchnorm}, \texttt{ReLU} \\
\texttt{conv3-3}, \texttt{tanh}, \texttt{batchnorm} \\
\bottomrule
\end{tabular}\vspace{-1em}
}
\end{table}

\textbf{Training}.
%The teacher models are trained using Nesterov accelerated gradient~\cite{nesterov1983method} with momentum $0.9$ for $200$ epochs. We use batch size $128$ and weight decaying $5\cdot10^{-4}$. The initial learning rate is set to be $0.1$, and we reduce it by multiplying $0.2$ at $60$, $120$, and $160$ epochs, respectively. Next, in the proposed data-free adversarial KD, 
For training generators in maximization, we use Adam optimizer~\cite{kingma2014adam} with momentum~$0.5$ and learning rate~$10^{-3}$. On the other hand, for training students in minimization, we use Nesterov accelerated gradient~\cite{nesterov1983method} with momentum~$0.9$ and learning rate~$0.1$. The learning rates are annealed by cosine decaying~\cite{loshchilov2016sgdr}. We adopt the vanilla KD for data-free KD from WRN40-2 to WRN16-1 on CIFAR-10. We use $50$ epochs in the warm-up stage and $200$ epochs for the main adversarial KD, where each epoch consists of $400$ batches of batch size $256$. In the other cases, we adopt variational information distillation (VID)~\cite{ahn2019variational} to match intermediate layer outputs, where we reduce the number of batches per epoch to $200$; VID is one of the state-of-the-art KD variants, and it yields better student accuracy with faster convergence. For the weighting factor~$\alpha$ in \eqref{sec:dfmc:minimaxkd:02}, we perform experiments on $\alpha\in\{10^{-3},10^{-2},10^{-1},1,10\}$ and choose the best results. The generator update interval~$m$ is set to be $10$ for wide residual networks and $1$ for the others. Except the results in Table~\ref{sec:exp:kd:tbl:03}, we use one generator and one student in our data-free KD, i.e., $G=S=1$ in Algorithm~\ref{sec:imple:alg:01}.

\subsection{Data-free model compression}\label{sec:exp:mc}

\begin{table*}[t]
\centering
\caption{Comparison of the proposed data-free adversarial KD scheme to the previous works.\label{sec:exp:kd:tbl:01}}\vspace{-.5em}
{\small
\begin{tabular}{cccccccccc}
\toprule
% & & && Ours & Micaelli et al.~\cite{micaelli2019zero} & Chen et al.~\cite{chen2019data} & Yin et al.~\cite{yin2019dreaming} \\
\multirow{2}{*}{\shortstack[c]{Original\\dataset}}
 & Teacher (\# params)
 & Student (\# params)
 & \multirow{3}{*}{\shortstack[c]{Teacher\\accuracy\\(\%)}}
 & \multicolumn{6}{c}{Student accuracy (\%)} \\
\cmidrule{5-10}
 & & & & \multicolumn{4}{c}{Data-free KD methods}
 & \multirow{2}{*}{\shortstack[c]{Training\\from scratch*}} & VID~\cite{ahn2019variational}* \\
 & & & & Ours & \cite{micaelli2019zero} & \cite{chen2019data} & \cite{yin2019dreaming} \\
\midrule
SVHN
 & WRN40-2 (2.2M)%(2.25M / 9.00M)
 & WRN16-1 (0.2M)%(0.18M / 0.71M)
 & 98.04  & \textbf{96.48} & 94.06 & N/A   & N/A   & 97.67 & 97.60 \\
\midrule
\multirow{5.5}{*}{CIFAR-10}
 & \multirow{3}{*}{WRN40-2 (2.2M)}%(2.25M / 9.00M)}
 & WRN16-1 (0.2M)%(0.18M / 0.71M)
 & \multirow{3}{*}{94.77}
          & \textbf{86.14} & 83.69 & N/A   & N/A   & 90.97 & 91.78 \\
 &
 & WRN40-1 (0.6M)%(0.57M / 2.27M)
 &
          & \textbf{91.69} & 86.60 & N/A   & N/A   & 93.35 & 93.67 \\
 &
 & WRN16-2 (0.7M)%(0.69M / 2.78M)
 &
          & \textbf{92.01} & 89.71 & N/A   & N/A   & 93.72 & 94.06 \\
\cmidrule{2-10}
 & VGG-11 (9.2M)%(9.23M / 36.94M)
 & ResNet-18 (11.2M)%(11.18M / 44.74M)
 & 92.37  & \textbf{90.84} & N/A   & N/A   & 90.36 & 94.56 & 91.47 \\
 & ResNet-34 (21.3M)%(21.30M / 85.21M)
 & ResNet-18 (11.2M)%(11.18M / 44.74M)
 & 95.11  & \textbf{94.61} & N/A   & 92.22 & 93.26 & 94.56 & 94.90 \\
\midrule
CIFAR-100
 & ResNet-34 (21.3M)%(21.35M / 85.40M)
 & ResNet-18 (11.2M)%(11.23M / 44.93M)
 & 78.34  & \textbf{77.01} & N/A   & 74.47 & N/A   & 77.32 & 77.77 \\
\midrule
Tiny-ImageNet
 & ResNet-34 (21.4M)%(21.40M / 85.60M)
 & ResNet-18 (11.3M)%(11.28M / 45.13M)
 & 66.34  & \textbf{63.73} & N/A   & N/A   & N/A   & 64.87 & 66.01 \\
\bottomrule
\multicolumn{10}{r}{* Used the original datasets.}
%\multicolumn{10}{l}{* \# FLOPs are computed by using \url{https://www.tensorflow.org/api_docs/python/tf/compat/v1/profiler/profile}.}
\end{tabular}\vspace{-.5em}
}
\end{table*}

We evaluate the performance of the proposed data-free model compression scheme on SVHN, CIFAR-10, CIFAR-100, and Tiny-ImageNet datasets for KD of residual networks (ResNets) and wide residual networks (WRNs). We summarize the main results in Table~\ref{sec:exp:kd:tbl:01}. We compare our scheme to the previous data-free KD methods in \cite{micaelli2019zero,chen2019data,yin2019dreaming} and show that we achieve the state-of-the-art data-free KD performance in all evaluation cases. We also obtain the student accuracy when students are trained with the original datasets from scratch and by using variational information distillation (VID) in \cite{ahn2019variational}. Table~\ref{sec:exp:kd:tbl:01} shows that the accuracy losses of our data-free KD method are marginal, compared to the cases of using the original datasets.

\begin{figure}[t]
\centering
\includegraphics[width=.95\columnwidth]{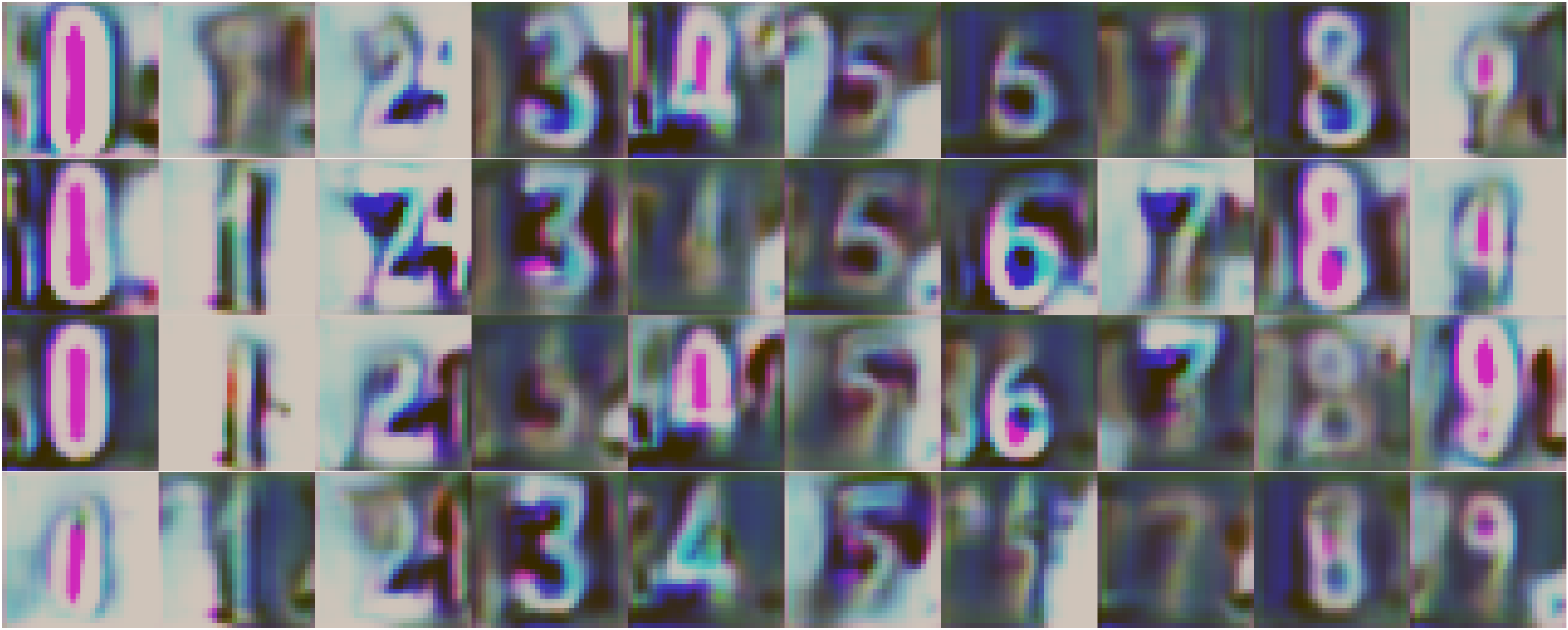}\vspace{-.5em}
\caption{Example synthetic images generated in data-free KD from WRN40-2 to WRN16-1 for SVHN. Just for better presentation, we classify the synthetic images using the teacher and show 4 samples from $0$ to $9$ in each column.
\label{sec:exp:kd:fig:01}}\vspace{-.5em}
\end{figure}
\begin{figure}[t]
\centering
\includegraphics[width=.95\columnwidth]{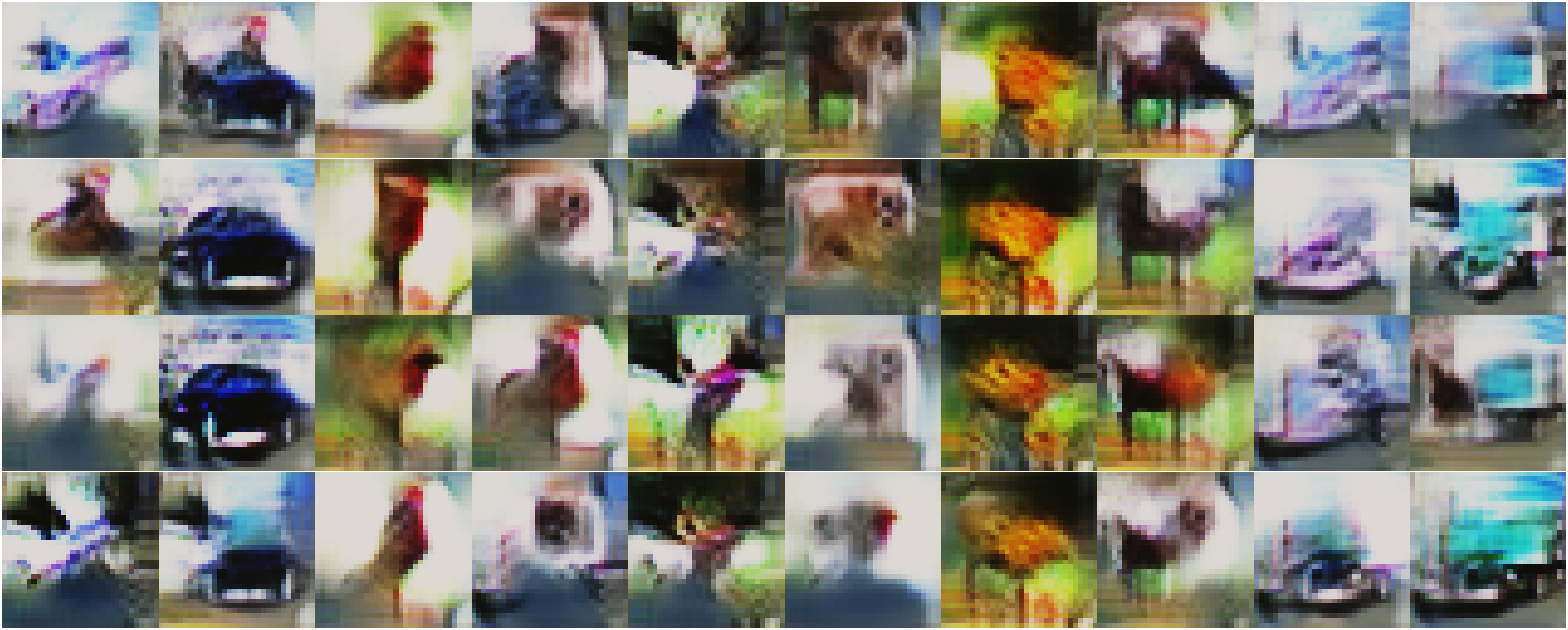}\vspace{-.5em}
\caption{Example synthetic images generated in data-free KD from WRN40-2 to WRN16-1 for CIFAR-10. Similar to Figure~\ref{sec:exp:kd:fig:01}, we classify the synthetic images using the teacher and show 4 samples for each class of CIFAR-10 (airplane, automobile, bird, cat, deer, dog, frog, horse, ship, and truck) in each column.
\label{sec:exp:kd:fig:02}}\vspace{-1em}
\end{figure}
\begin{figure}[t]
\centering
\includegraphics[width=.95\columnwidth]{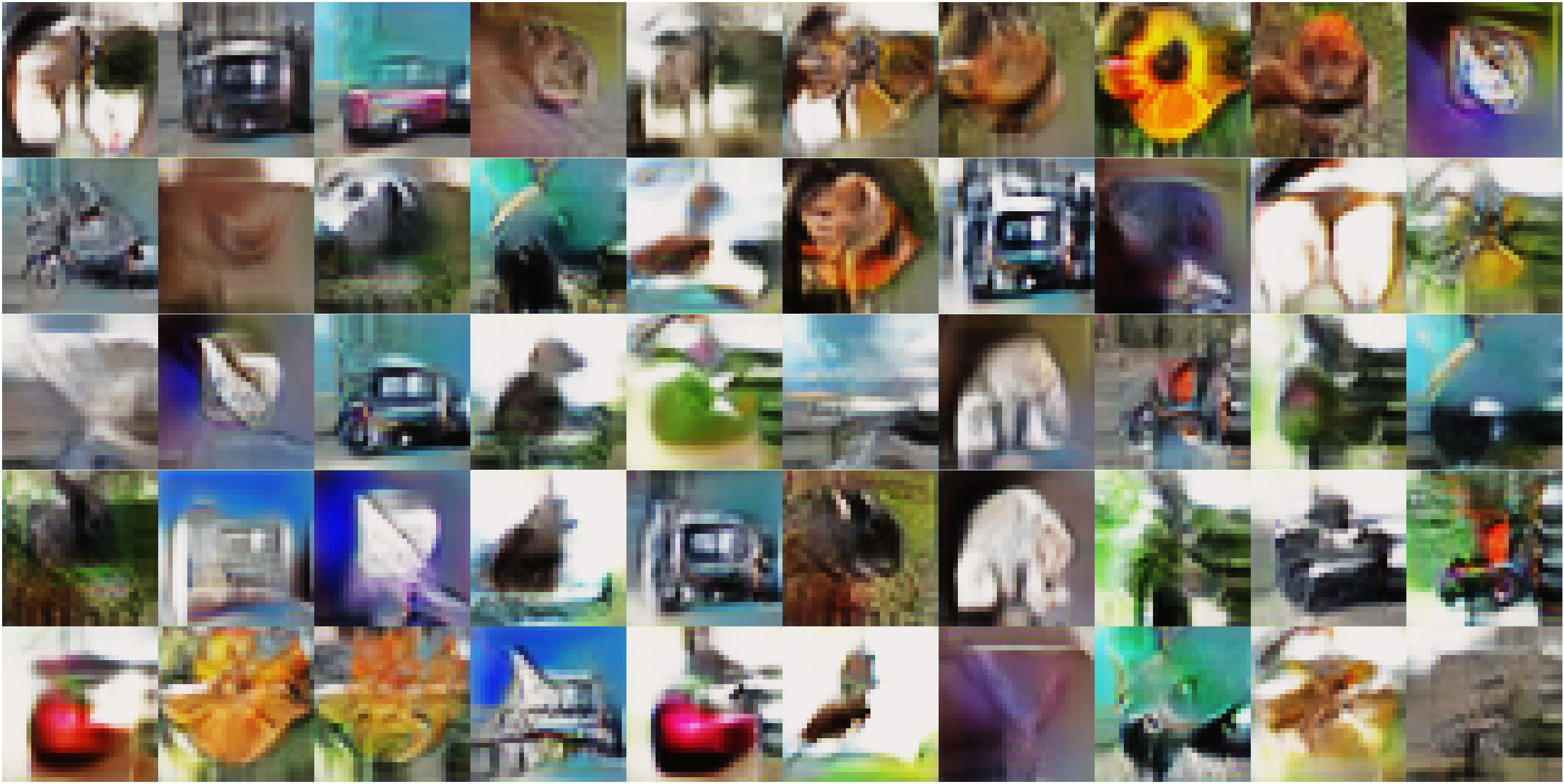}\vspace{-.5em}
\caption{Example synthetic images generated in data-free KD from ResNet-34 to ResNet-18 for CIFAR-100. %For simplicity, we show synthetic images without sorting them by class in this figure.
\label{sec:exp:kd:fig:03}}\vspace{-1em}
\end{figure}

\textbf{Example synthetic images}. We show example synthetic images obtained from generators trained with teachers pre-trained for SVHN, CIFAR-10, and CIFAR-100 datasets, respectively, in Figure~\ref{sec:exp:kd:fig:01}, Figure~\ref{sec:exp:kd:fig:02}, and Figure~\ref{sec:exp:kd:fig:03}. The figures show that the generators regularized with pre-trained teachers produce samples that are similar to the original datasets.

\begin{figure*}[t]
\centering
\includegraphics[width=.85\textwidth]{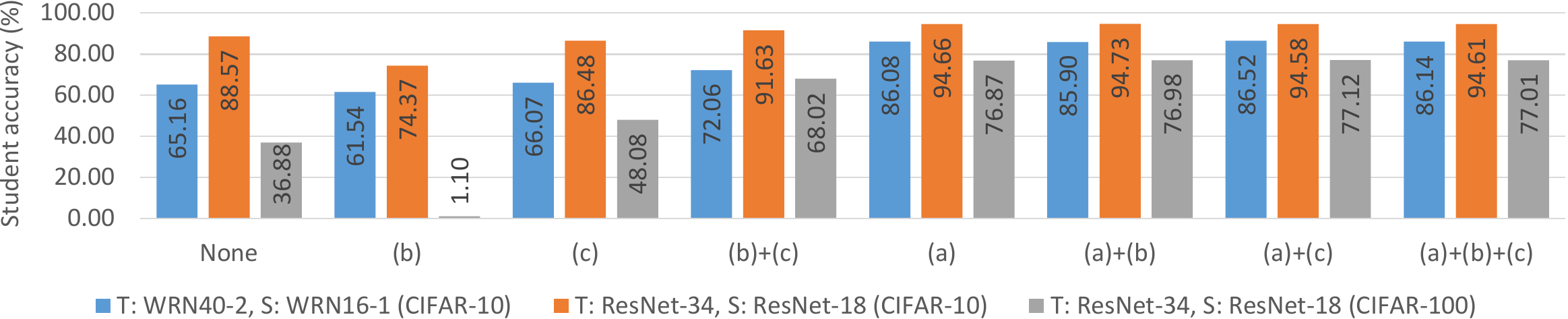}\vspace{-.5em}
\caption{Ablation study on the three terms in the auxiliary loss~$L_\psi$ of \eqref{sec:dfmc:const:01}, i.e., (a) batch normalization statistics, (b) instance categorical entropy, and (c) batch categorical entropy (see Section~\ref{sec:dfmc:const}). %Observe that the contribution from (a) batch normalization statistiscs is the largest.
\label{sec:exp:kd:fig:04}}\vspace{-1em}
\end{figure*}
\begin{figure}[t]
\centering
{\small
\begin{tabular}{ll}
%\includegraphics[height=.53\columnwidth]{figs/model0vidi_WRN40-2toWRN16-1cifar10_kl-crop.pdf}
% & \includegraphics[height=.53\columnwidth]{figs/model0vidi_WRN40-2toWRN16-1cifar10_acc-crop.pdf} \\
\includegraphics[height=.45\columnwidth]{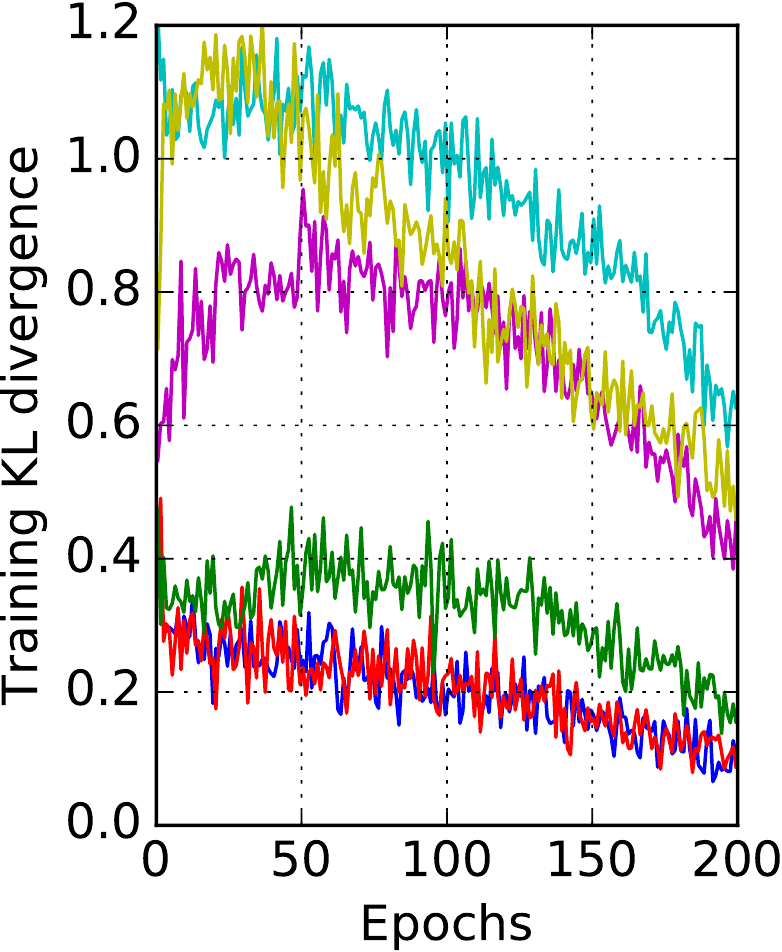}
 & \includegraphics[height=.45\columnwidth]{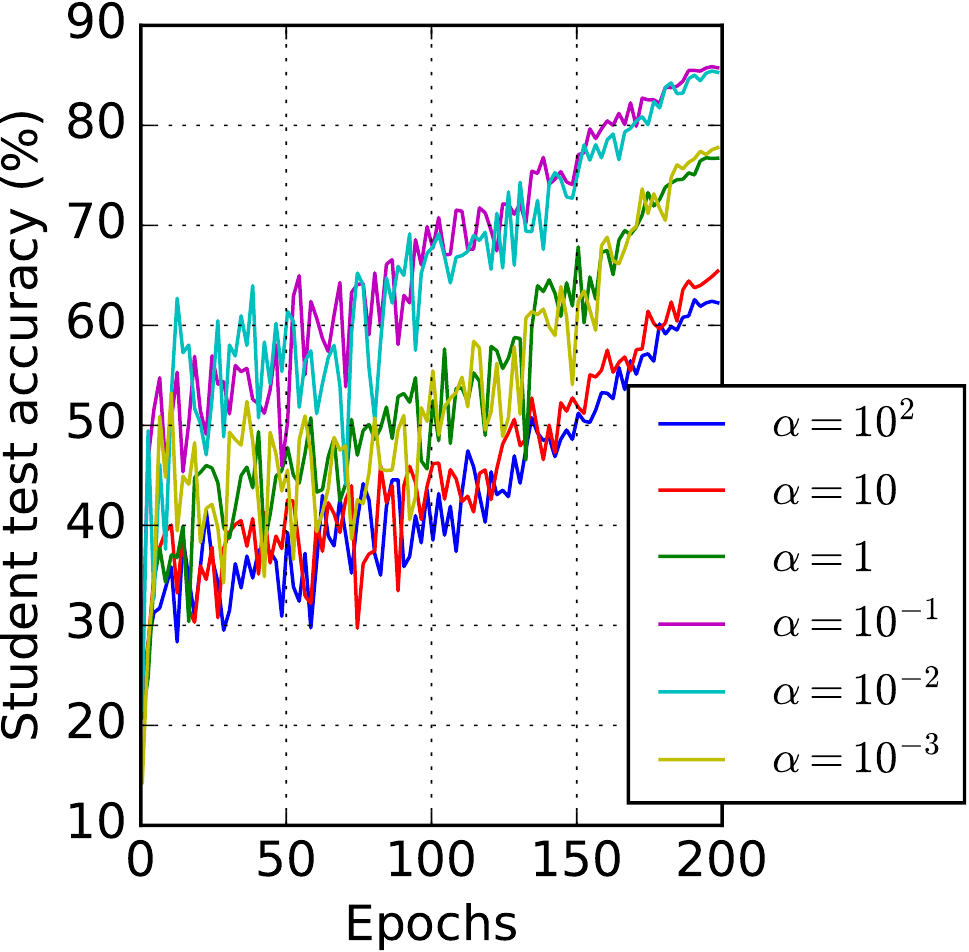} \\
(a) Training KL divergence
 &~~(b) Student test accuracy \\
\end{tabular}\vspace{-.5em}
}
\caption{Training KL divegence and student test accuracy of data-free KD for different values of $\alpha$ in \eqref{sec:dfmc:minimaxkd:02}. The student over-fits to the generator output when the weighting factor~$\alpha$ is too large ($\alpha\in\{10,10^2\}$).
\label{sec:exp:kd:fig:05}}\vspace{-.5em}
\end{figure}
\begin{figure*}[t]
\centering
{\small
\setlength\tabcolsep{0em}
\begin{tabular}{ccccccccc}
Epochs & ~\quad~ & Automobile & Bird & Horse & ~\quad~ & Automobile & Bird & Horse \\
%\cmidrule(l{4pt}r{4pt}){1-1}
\cmidrule(l{4pt}r{4pt}){3-5}
\cmidrule(l{4pt}r{4pt}){7-9}
\shortstack[c]{10\\~\\~\\~}
 & ~~ 
 & \includegraphics[width=.3\columnwidth]{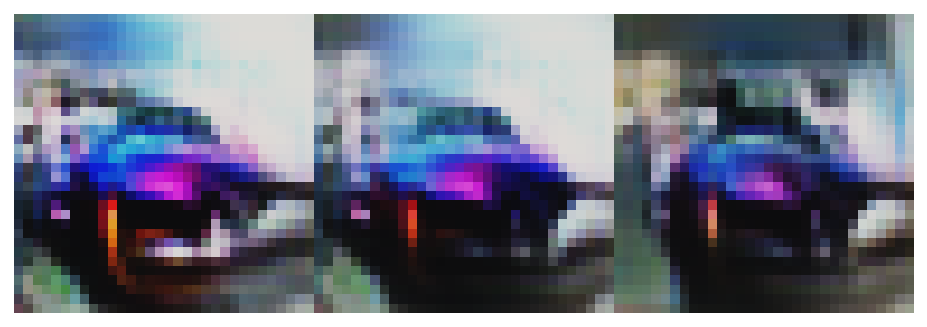}\vspace{-.1em}
 & \includegraphics[width=.3\columnwidth]{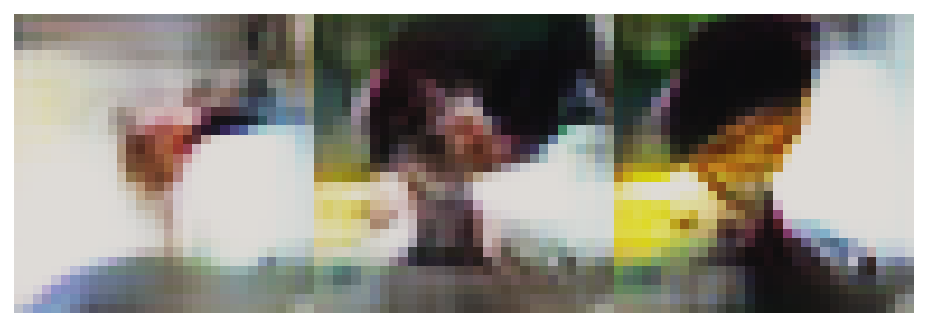}\vspace{-.1em}
 & \includegraphics[width=.3\columnwidth]{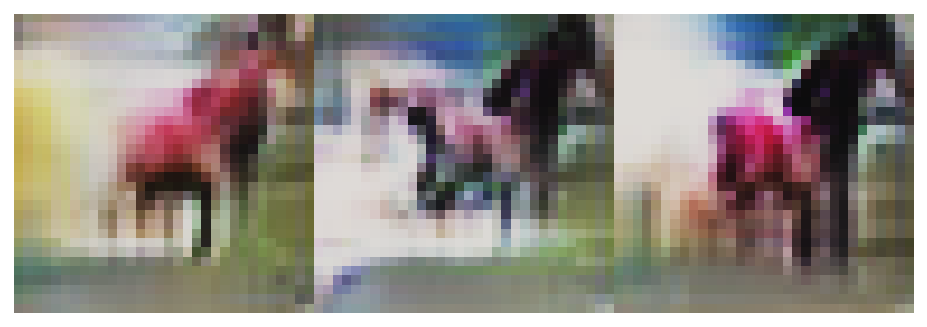}\vspace{-.1em}
 & ~~
 & \includegraphics[width=.3\columnwidth]{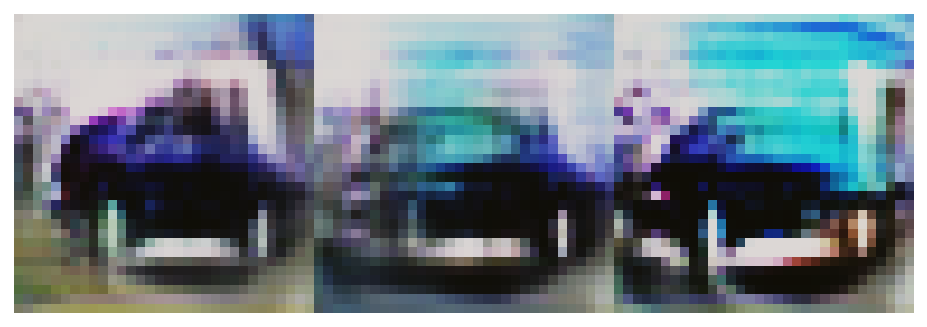}\vspace{-.1em}
 & \includegraphics[width=.3\columnwidth]{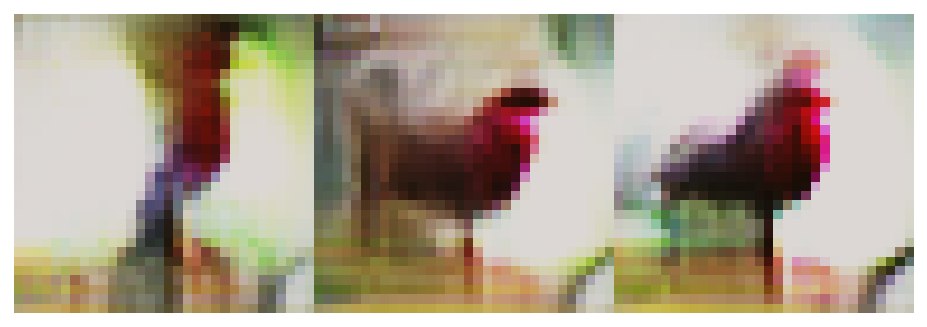}\vspace{-.1em}
 & \includegraphics[width=.3\columnwidth]{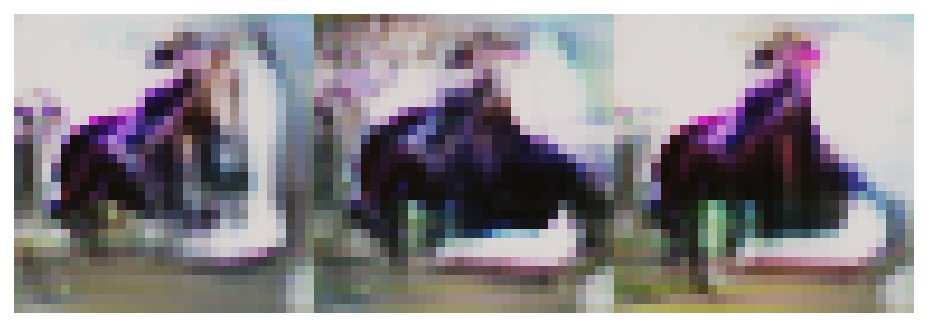}\vspace{-.1em} \\
\shortstack[c]{50\\~\\~\\~}
 & ~~ 
 & \includegraphics[width=.3\columnwidth]{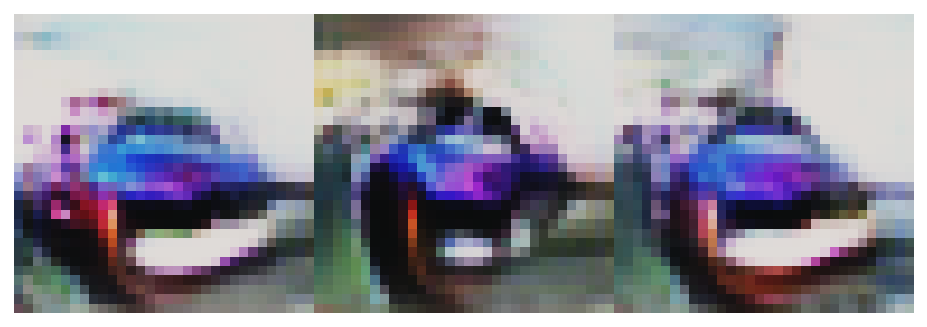}\vspace{-.1em}
 & \includegraphics[width=.3\columnwidth]{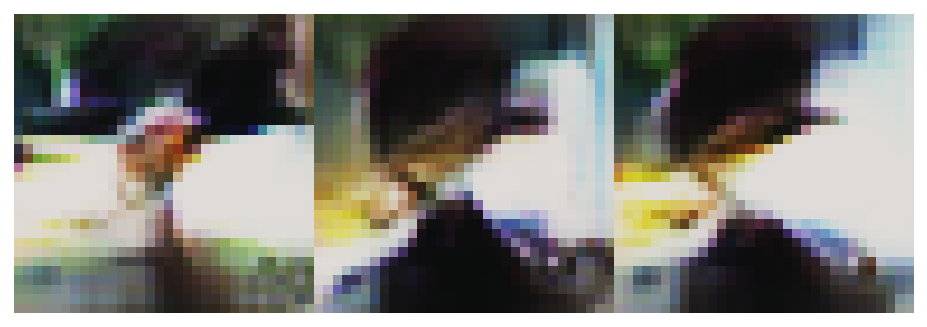}\vspace{-.1em}
 & \includegraphics[width=.3\columnwidth]{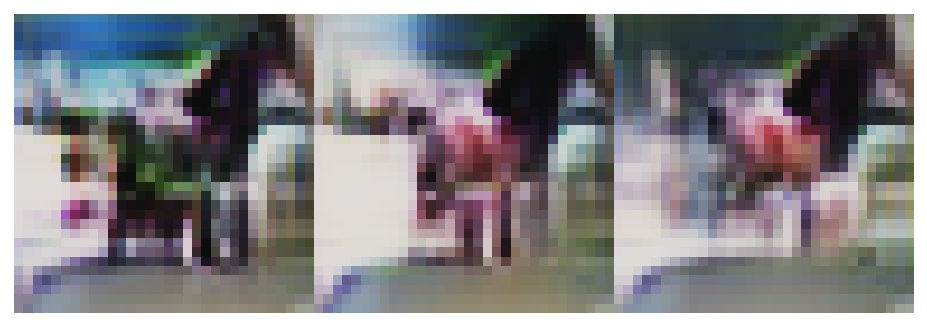}\vspace{-.1em}
 & ~~
 & \includegraphics[width=.3\columnwidth]{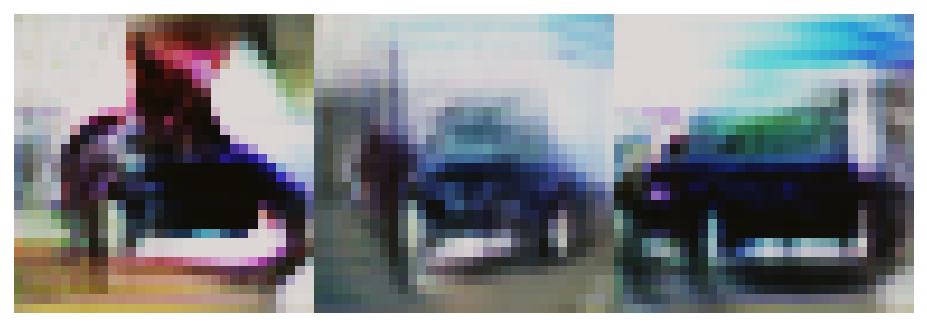}\vspace{-.1em}
 & \includegraphics[width=.3\columnwidth]{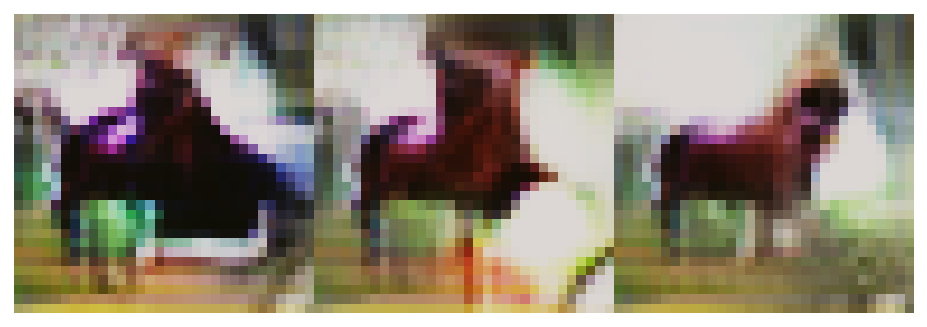}\vspace{-.1em}
 & \includegraphics[width=.3\columnwidth]{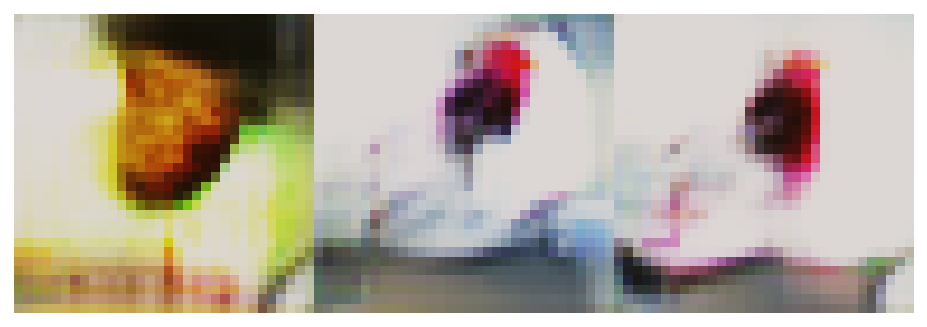}\vspace{-.1em} \\
\shortstack[c]{100\\~\\~\\~}
 & ~~ 
 & \includegraphics[width=.3\columnwidth]{figs/res/model0vidi_WRN40-2toWRN16-1cifar10data0_64C_512z_0cgan_1gan_1std_10gint_1_01_01_00_00_10_0lambda_0seed_1class_0gan_samples_0049epoch}\vspace{-.1em}
 & \includegraphics[width=.3\columnwidth]{figs/res/model0vidi_WRN40-2toWRN16-1cifar10data0_64C_512z_0cgan_1gan_1std_10gint_1_01_01_00_00_10_0lambda_0seed_2class_0gan_samples_0049epoch}\vspace{-.1em}
 & \includegraphics[width=.3\columnwidth]{figs/res/model0vidi_WRN40-2toWRN16-1cifar10data0_64C_512z_0cgan_1gan_1std_10gint_1_01_01_00_00_10_0lambda_0seed_7class_0gan_samples_0049epoch}\vspace{-.1em}
 & ~~
 & \includegraphics[width=.3\columnwidth]{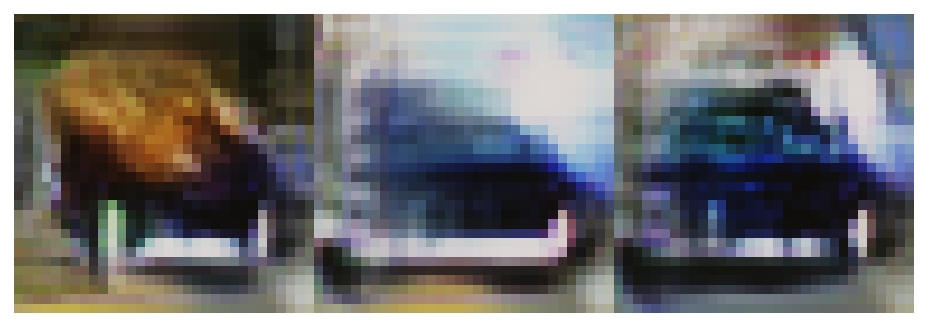}\vspace{-.1em}
 & \includegraphics[width=.3\columnwidth]{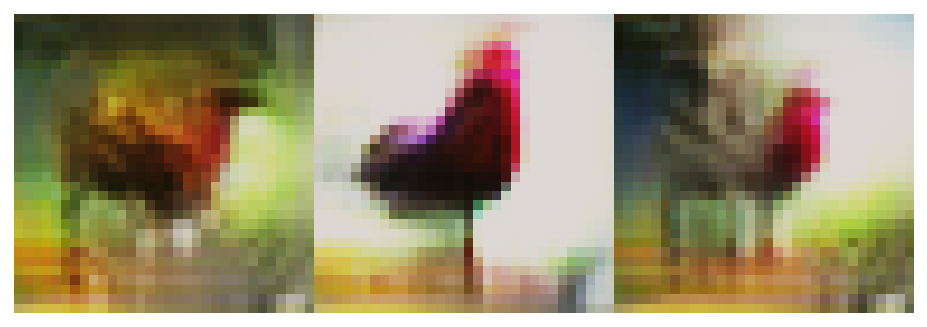}\vspace{-.1em}
 & \includegraphics[width=.3\columnwidth]{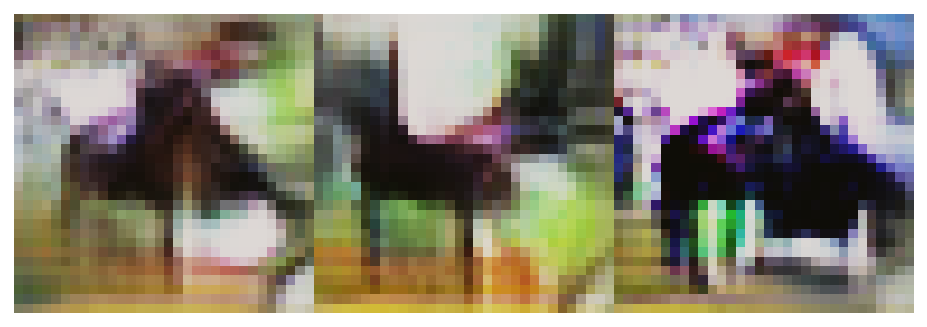}\vspace{-.1em} \\
%\shortstack[c]{150\\~\\~\\~}
% & ~~ 
% & \includegraphics[width=.3\columnwidth]{figs/res/model0vidi_WRN40-2toWRN16-1cifar10data0_64C_512z_0cgan_1gan_1std_10gint_1_01_01_00_00_10_0lambda_0seed_1class_0gan_samples_0149epoch}\vspace{-.1em}
% & \includegraphics[width=.3\columnwidth]{figs/res/model0vidi_WRN40-2toWRN16-1cifar10data0_64C_512z_0cgan_1gan_1std_10gint_1_01_01_00_00_10_0lambda_0seed_2class_0gan_samples_0149epoch}\vspace{-.1em}
% & \includegraphics[width=.3\columnwidth]{figs/res/model0vidi_WRN40-2toWRN16-1cifar10data0_64C_512z_0cgan_1gan_1std_10gint_1_01_01_00_00_10_0lambda_0seed_7class_0gan_samples_0149epoch}\vspace{-.1em}
% & ~~
% & \includegraphics[width=.3\columnwidth]{figs/res/model0vidi_WRN40-2toWRN16-1cifar10data0_64C_512z_0cgan_1gan_1std_10gint_1_01_01_00_01_00_0lambda_0seed_1class_0gan_samples_0149epoch}\vspace{-.1em}
% & \includegraphics[width=.3\columnwidth]{figs/res/model0vidi_WRN40-2toWRN16-1cifar10data0_64C_512z_0cgan_1gan_1std_10gint_1_01_01_00_01_00_0lambda_0seed_2class_0gan_samples_0149epoch}\vspace{-.1em}
% & \includegraphics[width=.3\columnwidth]{figs/res/model0vidi_WRN40-2toWRN16-1cifar10data0_64C_512z_0cgan_1gan_1std_10gint_1_01_01_00_01_00_0lambda_0seed_7class_0gan_samples_0149epoch}\vspace{-.1em} \\
\shortstack[c]{200\\~\\~\\~}
 & ~~ 
 & \includegraphics[width=.3\columnwidth]{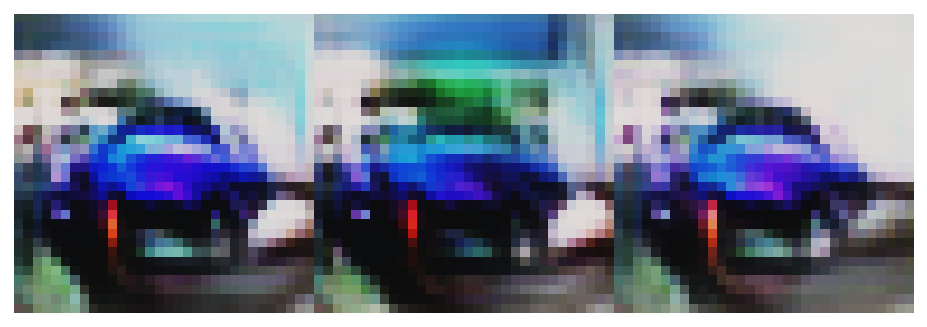}
 & \includegraphics[width=.3\columnwidth]{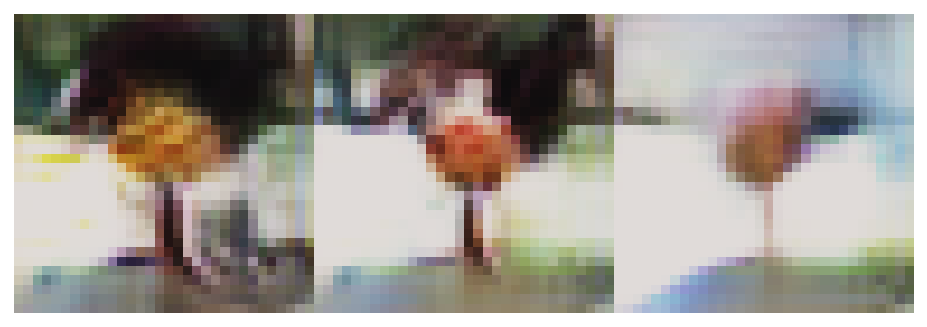}
 & \includegraphics[width=.3\columnwidth]{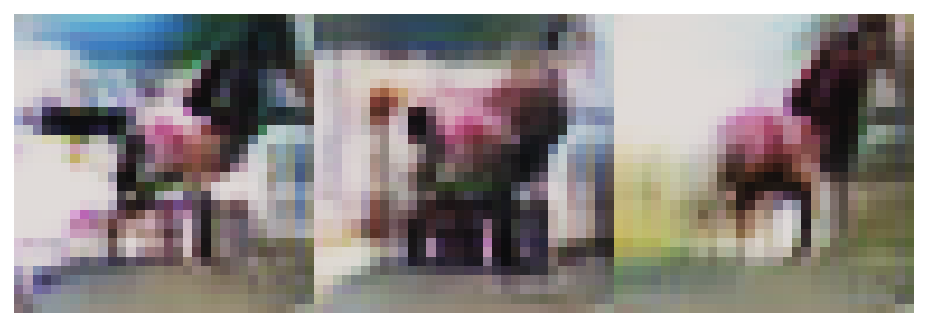}
 & ~~
 & \includegraphics[width=.3\columnwidth]{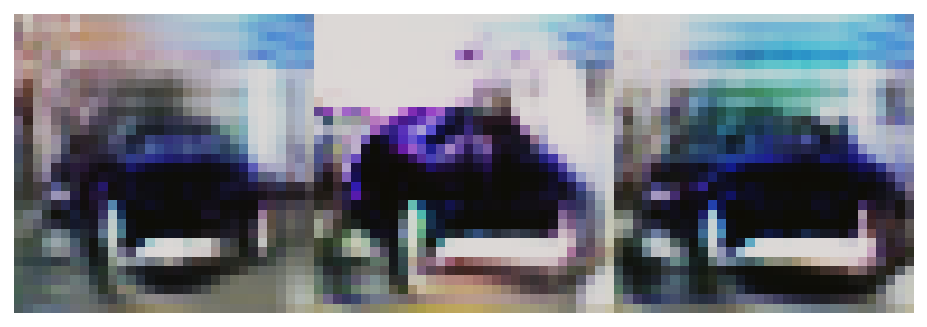}
 & \includegraphics[width=.3\columnwidth]{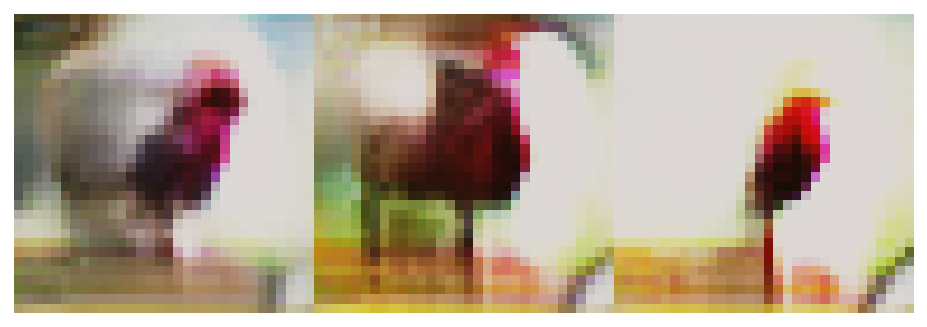}
 & \includegraphics[width=.3\columnwidth]{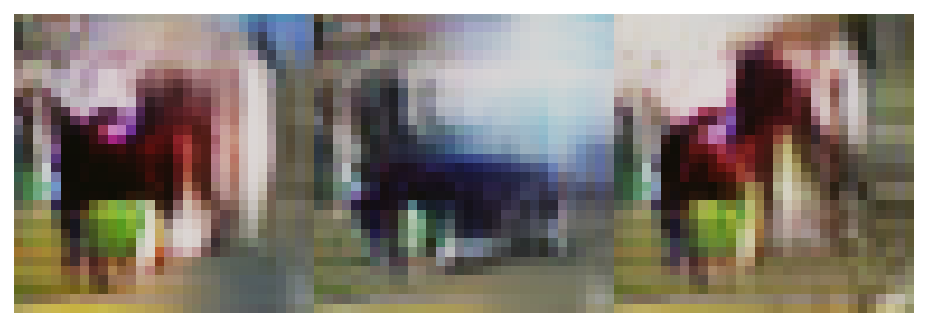} \\
 & ~~ & \multicolumn{3}{c}{(a) $\alpha=10$}  & ~~ & \multicolumn{3}{c}{(b) $\alpha=0.1$}
\end{tabular}\vspace{-.5em}
}
\caption{Example synthetic images generated for CIFAR-10 automobile, bird, and horse classes in different training epochs. We compare two cases with $\alpha=10$ and $\alpha=0.1$ to show the impact of the weighting factor~$\alpha$ in \eqref{sec:dfmc:minimaxkd:02} on the generator output.\label{sec:exp:kd:fig:06}}\vspace{-.5em}
\end{figure*}
%\begin{table}
%\centering
%\caption{Impact of using different datasets for KD. We denote the teacher and student networks after T: and S: in the Table, respectively.\label{sec:exp:kd:tbl:02}}\vspace{-.5em}
%{\small
%\begin{tabular}{cccc}
%\toprule
%%\backslashbox[0pt][l]{Data for KD}{Data for Inference} & SVHN & CIFAR-10 & CIFAR-100 \\
%\multirow{2}{*}{\shortstack[c]{Dataset used\\in KD}}
%          & \multicolumn{3}{c}{\shortstack[c]{Student accuracy (\%)}} \\
%\cmidrule{2-4}
%          & \shortstack[c]{T: WRN40-2\\S: WRN16-1\\(SVHN)}
%                  & \shortstack[c]{T: ResNet-34\\S: ResNet-18\\(CIFAR-10)}
%                          & \shortstack[c]{T: ResNet-34\\S: ResNet-18\\(CIFAR-100)} \\
%\midrule
%SVHN      & 97.60 & 92.89 & 50.81 \\
%CIFAR-10  & 95.56 & 94.90 & 77.25 \\
%CIFAR-100 & 95.91 & 94.81 & 77.77 \\
%\midrule
%Ours (data-free)
%          & 96.48 & 94.61 & 77.01 \\
%\bottomrule
%\end{tabular}\vspace{-1em}
%}
%\end{table}

\textbf{Ablation study}. For ablation study, we evaluate the proposed data-free KD scheme with and without each term of the auxiliary loss~$L_{\psi}$ for the generator in \eqref{sec:dfmc:const:01}, and the results are summarized in Figure~\ref{sec:exp:kd:fig:04}. The bar graph shows that the major contribution comes from (a), which is to match batch normalization statistics (see Section~\ref{sec:dfmc:const}). In Figure~\ref{sec:exp:kd:fig:05}, we present the impact of the weighting factor~$\alpha$ in \eqref{sec:dfmc:minimaxkd:02} on KD performance. Moreover, to visually show the impact of $\alpha$ on the generation of synthetic data, we collect synthetic images for $\alpha=10$ and $\alpha=0.1$ and show them at different epochs in Figure~\ref{sec:exp:kd:fig:06}. The figures show that smaller $\alpha$ yields more diverse adversarial images, since the generator is constrained less. As $\alpha$ gets larger, the generated images collapse to one mode for each class, which leads to over-fitting.
%We also compare our data-free scheme to the cases of using the alternative datasets. For alternative datasets, we consider two cases, (1) when a similar dataset is used instead of the original training dataset (e.g., CIFAR-100 instead of CIFAR-10) and (2) when a mismatched dataset is used instead (e.g., SVHN instead of CIFAR-10). The results in Table~\ref{sec:exp:kd:tbl:02} shows that using an alternative dataset similar to the original dataset works well and achieves similar performance to using the original dataset, while using a mismatched dataset, the performance degrades considerably. The proposed data-free scheme achieves comparable performance of using the original or alternative dataset similar to the original. We note that even obtaining alternative data is similarly difficult to obtaining the original data.

\begin{table}
\centering
\caption{Comparison of the student accuracy (\%) when using multiple generators and/or multiple students in our data-free KD from WRN40-2 to WRN16-1 on CIFAR-10.\label{sec:exp:kd:tbl:03}}\vspace{-.5em}
{\small
\begin{tabular}{c|cc}%c}
\toprule
\backslashbox[0pt][l]{\# students ($S$)}{\# generators ($G$)} & 1 & 2 \\%& 4 \\
\hline
1 & 86.14 & 86.67 \\
2 & 86.44 & \textbf{87.04} \\
\bottomrule
\end{tabular}\vspace{-1em}
}
\end{table}

\textbf{Multiple generators and multiple students}. We show the gain of using multiple generators and/or multiple students in Table~\ref{sec:exp:kd:tbl:03}. We compare the cases of using two generators and/or two students. For the second generator, we replace one middle convolutional layer with a residual block. For KD to two students, we use identical students with different initialization. Table~\ref{sec:exp:kd:tbl:03} shows that increasing the number of generators and/or the number of students results in better student accuracy in data-free KD.

\subsection{Data-free network quantization}\label{sec:exp:mq}

\begin{table*}[t]
\centering
\caption{Results of network quantization with the proposed data-free adversarial KD scheme. For our data-free quantization, we show the results for data-free quantization only (DF-Q) and data-free quantization-aware training with data-free KD (DF-QAT-KD). For conventional data-dependent quantization~\cite{jacob2018quantization}, we show the results for quantization only (Q), quantization-aware training (QAT), and quantization-aware training with KD (QAT-KD).\label{sec:exp:quant:tbl:01}}\vspace{-.5em}
{\small
\begin{tabular}{cccccccc}
\toprule
\multirow{2}{*}{\shortstack[c]{Original\\dataset}}
         & \multirow{2}{*}{\shortstack[c]{Pre-trained model\\(accuracy \%)}}
                      & \multirow{2}{*}{\shortstack[c]{Quantization bit-width\\for weights / activations}}
                              & \multicolumn{5}{c}{Quantized model accuracy (\%)} \\
\cmidrule{4-8}
         &            &       & \multicolumn{2}{c}{Ours (data-free)}
                                                          & \multicolumn{3}{c}{Data-dependent~\cite{jacob2018quantization}*} \\
         &            &       & DF-Q     & DF-QAT-KD      & Q & QAT & QAT-KD \\
\midrule
\multirow{2}{*}{SVHN}
         & \multirow{2}{*}{WRN16-1 (97.67)}
                      & 8 / 8 & 97.67    & 97.74          & 97.70 & 97.71    & 97.78 \\
         &            & 4 / 8 & 91.92    & 97.53          & 93.83 & 97.66    & 97.70 \\
%         &            & 4 / 4 & 82.83    & 94.50          & 81.34 & 97.08    & 97.06 \\
\midrule
\multirow{4.5}{*}{CIFAR-10}
         & \multirow{2}{*}{WRN16-1 (90.97)}
                      & 8 / 8 & 90.51    & 90.90          & 90.95 & 91.21    & 91.16 \\
         &            & 4 / 8 & 86.29    & 88.91          & 86.74 & 90.92    & 90.71 \\
\cmidrule{2-8}
         & \multirow{2}{*}{WRN40-2 (94.77)}
                      & 8 / 8 & 94.47    & 94.76          & 94.75 & 94.91    & 95.02 \\
         &            & 4 / 8 & 93.14    & 94.22          & 93.56 & 94.73    & 94.42 \\
\midrule
\multirow{2}{*}{CIFAR-100}
         & \multirow{2}{*}{ResNet-18 (77.32)}
                      & 8 / 8 & 76.68    & 77.30          & 77.43 & 77.84    & 77.73 \\
         &            & 4 / 8 & 71.02    & 75.15          & 69.63 & 75.52    & 75.62 \\
\midrule
Tiny-ImageNet
         & MobileNet v1 (64.34)
                      & 8 / 8 & 51.76    & 63.11          & 54.48 & 61.94    & 64.53 \\
\bottomrule
\multicolumn{8}{r}{* Used the original datasets.}
\end{tabular}\vspace{-1.5em}
}
\end{table*}
\begin{table}
\centering
\caption{Impact of using different datasets for 4-bit weight and 8-bit activation quantization.\label{sec:exp:quant:tbl:02}}\vspace{-.5em}
{\small
\begin{tabular}{cccc}
\toprule
\shortstack[c]{Dataset used\\in KD}
          & \multicolumn{3}{c}{\shortstack[c]{Quantized model accuracy (\%)\\before / after fine-tuning with KD}} \\
\cmidrule{2-4}
          & \shortstack[c]{WRN16-1\\(SVHN)}
                          & \shortstack[c]{WRN40-2\\(CIFAR-10)}
                                          & \shortstack[c]{ResNet-18\\(CIFAR-100)} \\
\midrule
SVHN      & 93.83 / 97.70 & 71.89 / 92.08 & 13.41 / 65.07 \\
CIFAR-10  & 93.50 / 97.24 & 93.56 / 94.42 & 67.50 / 75.62 \\
CIFAR-100 & 94.11 / 97.26 & 92.18 / 94.10 & 69.63 / 75.62 \\
\midrule
Ours (data-free)
          & 91.92 / 97.53 & 93.14 / 94.22 & 71.02 / 75.15 \\
\bottomrule
\end{tabular}\vspace{-1em}
}
\end{table}

In this subsection, we present the experimental results of the proposed data-free adversarial KD scheme on network quantization. For the baseline quantization scheme, we use TensorFlow's quantization framework. In particular, we implement our data-free KD scheme in the quantization-aware training framework~\cite{jacob2018quantization,krishnamoorthi2018quantizing} of TensorFlow\footnote{\url{https://github.com/tensorflow/tensorflow/tree/r1.15/tensorflow/contrib/quantize}}.

TensorFlow's quantization-aware training performs per-layer asymmetric quantization of weights and activations. For quantization only, no data are needed for weight quantization, but quantization of activations requires representative data, which are used to collect the range (the minimum and the maximum) of activations and to determine the quantization bin size based on the range. In our data-free quantization, we use synthetic data from a generator as the representative data. To this end, we train a generator with no adversarial loss as in the warm-up stage of Algorithm~\ref{sec:imple:alg:01} (see DF-Q in Table~\ref{sec:exp:quant:tbl:01}). For our data-free quantization-aware training, we utilize the proposed adversarial KD on top of Tensorflow's quantization-aware framework, where a quantized network is set as the student and a pre-trained floating-point model is given as the teacher, which is denoted by DF-QAT-KD in Table~\ref{sec:exp:quant:tbl:01}.

We follow the training hyperparameters as described in Section~\ref{sec:exp:mc}, while we set the initial learning rate for KD to be $10^{-3}$. We use $200$ epochs for the warm-up stage and $50$ epochs for quantization-aware training with data-free KD. We adopt the vanilla KD with no intermediate layer output matching terms. We summarize the results in Table~\ref{sec:exp:quant:tbl:01}.

For comparison, we evaluate three conventional data-dependent quantization schemes using the original training datasets, i.e., quantization only (Q), quantization-aware training (QAT), and quantization-aware training with KD (QAT-KD). As presented in Table~\ref{sec:exp:quant:tbl:01}, our data-free quantization shows very marginal accuracy losses less than 2\% for 4-bit/8-bit weight and 8-bit activation quantization in all the evaluated cases, compared to using the original datasets.

Finally, we compare our data-free quantization to using alternative datasets. We consider two cases (1) when a similar dataset is used (e.g., CIFAR-100 instead of CIFAR-10) and (2) when a mismatched dataset is used (e.g., SVHN instead of CIFAR-10). The results in Table~\ref{sec:exp:quant:tbl:02} show that using a mismatched dataset degrades the performance considerably. Using a similar dataset achieves comparable performance to our data-free scheme, which shows small accuracy losses less than 0.5\% compared to using the original datasets.
%The proposed data-free scheme achieves comparable performance to using the original dataset or an alternative dataset similar to the original dataset. However, we note that even obtaining alternative data, which are safe from any privacy and regulatory concerns, is difficult similarly to obtaining the original data in usual cases.
We note that even alternative data, which are safe from privacy and regulatory concerns, are hard to collect in usual cases.

\section{Conclusion}

In this paper, we proposed data-free adversarial KD for network quantization and compression. No original data are used in the proposed framework, while we train a generator to produce synthetic data adversarial to KD. In particular, we propose matching batch normalization statistics in the teacher to additionally constrain the generator to produce samples similar to the original training data. We used the proposed data-free KD scheme for compression of various models trained on SVHN, CIFAR-10, CIFAR-100, and Tiny-ImageNet datasets. In our experiments, we achieved the state-of-the-art data-free KD performance over the existing data-free KD schemes. For network quantization, we obtained quantized models that achieve comparable accuracy to the models quantized and fine-tuned with the original training datasets. The proposed framework shows great potential to keep data privacy in model compression.

{\small
\bibliographystyle{ieee_fullname}
\bibliography{ref}

\begin{thebibliography}{10}\itemsep=-1pt

\bibitem{lecun2015deep}
Yann LeCun, Yoshua Bengio, and Geoffrey Hinton.
\newblock Deep learning.
\newblock {\em Nature}, 521(7553):436--444, 2015.

\bibitem{goodfellow2016deep}
Ian Goodfellow, Yoshua Bengio, and Aaron Courville.
\newblock {\em Deep learning}.
\newblock MIT press, 2016.

\bibitem{sze2017efficient}
Vivienne Sze, Yu-Hsin Chen, Tien-Ju Yang, and Joel~S Emer.
\newblock Efficient processing of deep neural networks: A tutorial and survey.
\newblock {\em Proceedings of the IEEE}, 105(12):2295--2329, 2017.

\bibitem{cheng2018model}
Yu Cheng, Duo Wang, Pan Zhou, and Tao Zhang.
\newblock Model compression and acceleration for deep neural networks: The
  principles, progress, and challenges.
\newblock {\em IEEE Signal Processing Magazine}, 35(1):126--136, 2018.

\bibitem{han2015learning}
Song Han, Jeff Pool, John Tran, and William Dally.
\newblock Learning both weights and connections for efficient neural network.
\newblock In {\em Advances in Neural Information Processing Systems}, pages
  1135--1143, 2015.

\bibitem{wen2016learning}
Wei Wen, Chunpeng Wu, Yandan Wang, Yiran Chen, and Hai Li.
\newblock Learning structured sparsity in deep neural networks.
\newblock In {\em Advances in Neural Information Processing Systems}, pages
  2074--2082, 2016.

\bibitem{guo2016dynamic}
Yiwen Guo, Anbang Yao, and Yurong Chen.
\newblock Dynamic network surgery for efficient {DNNs}.
\newblock In {\em Advances In Neural Information Processing Systems}, pages
  1379--1387, 2016.

\bibitem{molchanov2017variational}
Dmitry Molchanov, Arsenii Ashukha, and Dmitry Vetrov.
\newblock Variational dropout sparsifies deep neural networks.
\newblock In {\em International Conference on Machine Learning}, pages
  2498--2507, 2017.

\bibitem{louizos2017bayesian}
Christos Louizos, Karen Ullrich, and Max Welling.
\newblock Bayesian compression for deep learning.
\newblock In {\em Advances in Neural Information Processing Systems}, pages
  3290--3300, 2017.

\bibitem{louizos2018learning}
Christos Louizos, Max Welling, and Diederik~P Kingma.
\newblock Learning sparse neural networks through {$L_0$} regularization.
\newblock In {\em International Conference on Learning Representations}, 2018.

\bibitem{frankle2018lottery}
Jonathan Frankle and Michael Carbin.
\newblock The lottery ticket hypothesis: Finding sparse, trainable neural
  networks.
\newblock In {\em International Conference on Learning Representations}, 2018.

\bibitem{dai2018compressing}
Bin Dai, Chen Zhu, Baining Guo, and David Wipf.
\newblock Compressing neural networks using the variational information
  bottleneck.
\newblock In {\em International Conference on Machine Learning}, pages
  1135--1144, 2018.

\bibitem{han2015deep}
Song Han, Huizi Mao, and William~J Dally.
\newblock Deep compression: Compressing deep neural networks with pruning,
  trained quantization and {Huffman} coding.
\newblock In {\em International Conference on Learning Representations}, 2016.

\bibitem{choi2017towards}
Yoojin Choi, Mostafa El-Khamy, and Jungwon Lee.
\newblock Towards the limit of network quantization.
\newblock In {\em International Conference on Learning Representations}, 2017.

\bibitem{ullrich2017soft}
Karen Ullrich, Edward Meeds, and Max Welling.
\newblock Soft weight-sharing for neural network compression.
\newblock In {\em International Conference on Learning Representations}, 2017.

\bibitem{park2017weighted}
Eunhyeok Park, Junwhan Ahn, and Sungjoo Yoo.
\newblock Weighted-entropy-based quantization for deep neural networks.
\newblock In {\em Proceedings of the IEEE International Conference on Computer
  Vision}, pages 7197--7205, 2017.

\bibitem{tung2018deep}
Frederick Tung and Greg Mori.
\newblock Deep neural network compression by in-parallel pruning-quantization.
\newblock {\em IEEE Transactions on Pattern Analysis and Machine Intelligence},
  2018.

\bibitem{choi2020universal}
Yoojin Choi, Mostafa El-Khamy, and Jungwon Lee.
\newblock Universal deep neural network compression.
\newblock {\em IEEE Journal of Selected Topics in Signal Processing}, 2020.

\bibitem{rastegari2016xnor}
Mohammad Rastegari, Vicente Ordonez, Joseph Redmon, and Ali Farhadi.
\newblock {XNOR-Net}: Imagenet classification using binary convolutional neural
  networks.
\newblock In {\em Proceedings of the European Conference on Computer Vision},
  pages 525--542, 2016.

\bibitem{zhou2016dorefa}
Shuchang Zhou, Yuxin Wu, Zekun Ni, Xinyu Zhou, He Wen, and Yuheng Zou.
\newblock {DoReFa-Net}: Training low bitwidth convolutional neural networks
  with low bitwidth gradients.
\newblock {\em arXiv preprint arXiv:1606.06160}, 2016.

\bibitem{zhu2017trained}
Chenzhuo Zhu, Song Han, Huizi Mao, and William~J Dally.
\newblock Trained ternary quantization.
\newblock In {\em International Conference on Learning Representations}, 2017.

\bibitem{cai2017deep}
Zhaowei Cai, Xiaodong He, Jian Sun, and Nuno Vasconcelos.
\newblock Deep learning with low precision by half-wave {Gaussian}
  quantization.
\newblock In {\em Proceedings of the IEEE Conference on Computer Vision and
  Pattern Recognition}, pages 5918--5926, 2017.

\bibitem{zhang2018lq}
Dongqing Zhang, Jiaolong Yang, Dongqiangzi Ye, and Gang Hua.
\newblock {LQ-Nets}: Learned quantization for highly accurate and compact deep
  neural networks.
\newblock In {\em Proceedings of the European Conference on Computer Vision},
  pages 365--382, 2018.

\bibitem{jacob2018quantization}
Benoit Jacob, Skirmantas Kligys, Bo Chen, Menglong Zhu, Matthew Tang, Andrew
  Howard, Hartwig Adam, and Dmitry Kalenichenko.
\newblock Quantization and training of neural networks for efficient
  integer-arithmetic-only inference.
\newblock In {\em Proceedings of the IEEE Conference on Computer Vision and
  Pattern Recognition}, pages 2704--2713, 2018.

\bibitem{wang2019haq}
Kuan Wang, Zhijian Liu, Yujun Lin, Ji Lin, and Song Han.
\newblock {HAQ}: Hardware-aware automated quantization with mixed precision.
\newblock In {\em Proceedings of the IEEE Conference on Computer Vision and
  Pattern Recognition}, pages 8612--8620, 2019.

\bibitem{howard2017mobilenets}
Andrew~G Howard, Menglong Zhu, Bo Chen, Dmitry Kalenichenko, Weijun Wang,
  Tobias Weyand, Marco Andreetto, and Hartwig Adam.
\newblock {MobileNets}: Efficient convolutional neural networks for mobile
  vision applications.
\newblock {\em arXiv preprint arXiv:1704.04861}, 2017.

\bibitem{hinton2015distilling}
Geoffrey Hinton, Oriol Vinyals, and Jeff Dean.
\newblock Distilling the knowledge in a neural network.
\newblock {\em arXiv preprint arXiv:1503.02531}, 2015.

\bibitem{romero2015fitnets}
Adriana Romero, Nicolas Ballas, Samira~Ebrahimi Kahou, Antoine Chassang, Carlo
  Gatta, and Yoshua Bengio.
\newblock {FitNets}: Hints for thin deep nets.
\newblock In {\em International Conference on Learning Representations}, 2015.

\bibitem{zagoruyko2017paying}
Sergey Zagoruyko and Nikos Komodakis.
\newblock Paying more attention to attention: Improving the performance of
  convolutional neural networks via attention transfer.
\newblock In {\em International Conference on Learning Representations}, 2017.

\bibitem{ahn2019variational}
Sungsoo Ahn, Shell~Xu Hu, Andreas Damianou, Neil~D Lawrence, and Zhenwen Dai.
\newblock Variational information distillation for knowledge transfer.
\newblock In {\em Proceedings of the IEEE Conference on Computer Vision and
  Pattern Recognition}, pages 9163--9171, 2019.

\bibitem{chen2018darkrank}
Yuntao Chen, Naiyan Wang, and Zhaoxiang Zhang.
\newblock {DarkRank}: Accelerating deep metric learning via cross sample
  similarities transfer.
\newblock In {\em Proceedings of the AAAI Conference on Artificial
  Intelligence}, 2018.

\bibitem{park2019relational}
Wonpyo Park, Dongju Kim, Yan Lu, and Minsu Cho.
\newblock Relational knowledge distillation.
\newblock In {\em Proceedings of the IEEE Conference on Computer Vision and
  Pattern Recognition}, pages 3967--3976, 2019.

\bibitem{lopes2017data}
Raphael~Gontijo Lopes, Stefano Fenu, and Thad Starner.
\newblock Data-free knowledge distillation for deep neural networks.
\newblock In {\em NeurIPS Workshop on Learning with Limited Data}, 2017.

\bibitem{bhardwaj2019dream}
Kartikeya Bhardwaj, Naveen Suda, and Radu Marculescu.
\newblock Dream distillation: A data-independent model compression framework.
\newblock In {\em ICML Joint Workshop on On-Device Machine Learning and Compact
  Deep Neural Network Representations (ODML-CDNNR)}, 2019.

\bibitem{chen2019data}
Hanting Chen, Yunhe Wang, Chang Xu, Zhaohui Yang, Chuanjian Liu, Boxin Shi,
  Chunjing Xu, Chao Xu, and Qi Tian.
\newblock Data-free learning of student networks.
\newblock In {\em Proceedings of the IEEE International Conference on Computer
  Vision}, pages 3514--3522, 2019.

\bibitem{micaelli2019zero}
Paul Micaelli and Amos~J Storkey.
\newblock Zero-shot knowledge transfer via adversarial belief matching.
\newblock In {\em Advances in Neural Information Processing Systems}, pages
  9547--9557, 2019.

\bibitem{he2016deep}
Kaiming He, Xiangyu Zhang, Shaoqing Ren, and Jian Sun.
\newblock Deep residual learning for image recognition.
\newblock In {\em Proceedings of the IEEE Conference on Computer Vision and
  Pattern Recognition}, pages 770--778, 2016.

\bibitem{zagoruyko2016wide}
Sergey Zagoruyko and Nikos Komodakis.
\newblock Wide residual networks.
\newblock In {\em Proceedings of the British Machine Vision Conference}, pages
  87.1--87.12, 2016.

\bibitem{netzer2011reading}
Yuval Netzer, Tao Wang, Adam Coates, Alessandro Bissacco, Bo Wu, and Andrew~Y
  Ng.
\newblock Reading digits in natural images with unsupervised feature learning.
\newblock In {\em NeurIPS Workshop on Deep Learning and Unsupervised Feature
  Learning}, 2011.

\bibitem{krizhevsky2009learning}
Alex Krizhevsky.
\newblock Learning multiple layers of features from tiny images.
\newblock {\em Technical report, Univ. of Toronto}, 2009.

\bibitem{yin2019dreaming}
Hongxu Yin, Pavlo Molchanov, Zhizhong Li, Jose~M Alvarez, Arun Mallya, Derek
  Hoiem, Niraj~K Jha, and Jan Kautz.
\newblock Dreaming to distill: Data-free knowledge transfer via
  {DeepInversion}.
\newblock {\em arXiv preprint arXiv:1912.08795}, 2019.

\bibitem{krishnamoorthi2018quantizing}
Raghuraman Krishnamoorthi.
\newblock Quantizing deep convolutional networks for efficient inference: A
  whitepaper.
\newblock {\em arXiv preprint arXiv:1806.08342}, 2018.

\bibitem{nayak2019zero}
Gaurav~Kumar Nayak, Konda~Reddy Mopuri, Vaisakh Shaj, Venkatesh~Babu
  Radhakrishnan, and Anirban Chakraborty.
\newblock Zero-shot knowledge distillation in deep networks.
\newblock In {\em International Conference on Machine Learning}, pages
  4743--4751, 2019.

\bibitem{yoo2019knowledge}
Jaemin Yoo, Minyong Cho, Taebum Kim, and U Kang.
\newblock Knowledge extraction with no observable data.
\newblock In {\em Advances in Neural Information Processing Systems}, pages
  2701--2710, 2019.

\bibitem{nagel2019data}
Markus Nagel, Mart~van Baalen, Tijmen Blankevoort, and Max Welling.
\newblock Data-free quantization through weight equalization and bias
  correction.
\newblock In {\em Proceedings of the IEEE International Conference on Computer
  Vision}, pages 1325--1334, 2019.

\bibitem{ben2009robust}
Aharon Ben-Tal, Laurent El~Ghaoui, and Arkadi Nemirovski.
\newblock {\em Robust Optimization}, volume~28.
\newblock Princeton University Press, 2009.

\bibitem{bertsimas2011theory}
Dimitris Bertsimas, David~B Brown, and Constantine Caramanis.
\newblock Theory and applications of robust optimization.
\newblock {\em SIAM review}, 53(3):464--501, 2011.

\bibitem{akhtar2018threat}
Naveed Akhtar and Ajmal Mian.
\newblock Threat of adversarial attacks on deep learning in computer vision: A
  survey.
\newblock {\em IEEE Access}, 6:14410--14430, 2018.

\bibitem{goodfellow2015explaining}
Ian~J Goodfellow, Jonathon Shlens, and Christian Szegedy.
\newblock Explaining and harnessing adversarial examples.
\newblock In {\em International Conference on Learning Representations}, 2014.

\bibitem{carlini2017towards}
Nicholas Carlini and David Wagner.
\newblock Towards evaluating the robustness of neural networks.
\newblock In {\em IEEE Symposium on Security and Privacy}, pages 39--57, 2017.

\bibitem{madry2018towards}
Aleksander Madry, Aleksandar Makelov, Ludwig Schmidt, Dimitris Tsipras, and
  Adrian Vladu.
\newblock Towards deep learning models resistant to adversarial attacks.
\newblock In {\em International Conference on Learning Representations}, 2018.

\bibitem{poursaeed2018generative}
Omid Poursaeed, Isay Katsman, Bicheng Gao, and Serge Belongie.
\newblock Generative adversarial perturbations.
\newblock In {\em Proceedings of the IEEE Conference on Computer Vision and
  Pattern Recognition}, pages 4422--4431, 2018.

\bibitem{wang2019direct}
Huaxia Wang and Chun-Nam Yu.
\newblock A direct approach to robust deep learning using adversarial networks.
\newblock In {\em International Conference on Learning Representations}, 2019.

\bibitem{jang2019adversarial}
Yunseok Jang, Tianchen Zhao, Seunghoon Hong, and Honglak Lee.
\newblock Adversarial defense via learning to generate diverse attacks.
\newblock In {\em Proceedings of the IEEE International Conference on Computer
  Vision}, pages 2740--2749, 2019.

\bibitem{ulyanov2018deep}
Dmitry Ulyanov, Andrea Vedaldi, and Victor Lempitsky.
\newblock Deep image prior.
\newblock In {\em Proceedings of the IEEE Conference on Computer Vision and
  Pattern Recognition}, pages 9446--9454, 2018.

\bibitem{mordvintsev2015inceptionism}
Alexander Mordvintsev, Christopher Olah, and Mike Tyka.
\newblock Inceptionism: Going deeper into neural networks, 2015.
\newblock
  \url{https://research.googleblog.com/2015/06/inceptionism-going-deeper-into-neural.html}
  [Online; accessed 18-April-2020].

\bibitem{goodfellow2014generative}
Ian Goodfellow, Jean Pouget-Abadie, Mehdi Mirza, Bing Xu, David Warde-Farley,
  Sherjil Ozair, Aaron Courville, and Yoshua Bengio.
\newblock Generative adversarial nets.
\newblock In {\em Advances in Neural Information Processing Systems}, pages
  2672--2680, 2014.

\bibitem{goodfellow2016nips}
Ian Goodfellow.
\newblock {NIPS} 2016 tutorial: Generative adversarial networks.
\newblock {\em arXiv preprint arXiv:1701.00160}, 2016.

\bibitem{durugkar2017generative}
Ishan Durugkar, Ian Gemp, and Sridhar Mahadevan.
\newblock Generative multi-adversarial networks.
\newblock In {\em International Conference on Learning Representations}, 2017.

\bibitem{nguyen2017dual}
Tu Nguyen, Trung Le, Hung Vu, and Dinh Phung.
\newblock Dual discriminator generative adversarial nets.
\newblock In {\em Advances in Neural Information Processing Systems}, pages
  2670--2680, 2017.

\bibitem{arora2017generalization}
Sanjeev Arora, Rong Ge, Yingyu Liang, Tengyu Ma, and Yi Zhang.
\newblock Generalization and equilibrium in generative adversarial nets
  ({GANs}).
\newblock In {\em International Conference on Machine Learning}, pages
  224--232, 2017.

\bibitem{hoang2018mgan}
Quan Hoang, Tu~Dinh Nguyen, Trung Le, and Dinh Phung.
\newblock {MGAN}: Training generative adversarial nets with multiple
  generators.
\newblock In {\em International Conference on Learning Representations}, 2018.

\bibitem{cover2012elements}
Thomas~M Cover and Joy~A Thomas.
\newblock {\em Elements of Information Theory}.
\newblock John Wiley \& Sons, 2012.

\bibitem{kingma2014adam}
Diederik Kingma and Jimmy Ba.
\newblock Adam: A method for stochastic optimization.
\newblock In {\em International Conference on Learning Representations}, 2015.

\bibitem{nesterov1983method}
Yurii Nesterov.
\newblock A method for unconstrained convex minimization problem with the rate
  of convergence {$O(1/k^2)$}.
\newblock In {\em Doklady AN USSR}, volume 269, pages 543--547, 1983.

\bibitem{loshchilov2016sgdr}
Ilya Loshchilov and Frank Hutter.
\newblock {SGDR}: Stochastic gradient descent with warm restarts.
\newblock In {\em International Conference on Learning Representations}, 2017.

\end{thebibliography}
}

\end{document}